\documentclass{article}

\usepackage{microtype}
\usepackage{graphicx}
\usepackage{subcaption}
\usepackage{booktabs} 

\usepackage{hyperref}
\usepackage{xcolor}
\usepackage{multirow}
\usepackage[T1]{fontenc}
\usepackage{colortbl}
\usepackage{pifont}
\usepackage{amsmath}
\usepackage{amsfonts}
\definecolor{darkgreen}{RGB}{50,100,0}
\definecolor{darkred}{RGB}{200, 0, 0}
\definecolor{lightred}{RGB}{250, 200, 200}
\definecolor{lightblue}{RGB}{210, 220, 250}
\definecolor{dodgerblue}{RGB}{30,144,255}
\definecolor{select}{RGB}{222, 235, 247}
\definecolor{unselect}{RGB}{247, 207, 206}

\newcommand{\Benchmark}{PaperArena}
\newcommand{\Platform}{PaperArena-Hub}
\newcommand{\cmark}{\ding{51}}
\newcommand{\xmark}{\ding{55}}

\usepackage[preprint]{icml2026}

\usepackage{amsmath}
\usepackage{amssymb}
\usepackage{mathtools}
\usepackage{amsthm}

\usepackage[capitalize,noabbrev]{cleveref}

\theoremstyle{plain}

\theoremstyle{definition}

\theoremstyle{remark}

\usepackage[textsize=tiny]{todonotes}
\usepackage[most]{tcolorbox}
\usepackage{wrapfig}
\usepackage{makecell}

\definecolor{frameorange}{RGB}{218, 114, 27}
\definecolor{bgorange}{RGB}{253, 243, 235}

\definecolor{frameblue}{RGB}{0, 85, 150}
\definecolor{bgblue}{RGB}{235, 245, 255}

\definecolor{framegreen}{RGB}{34, 139, 34}
\definecolor{bggreen}{RGB}{240, 253, 240}

\newtcolorbox{StrategyBox}[3][frameorange]{
  enhanced,
  float*,
  width=\textwidth,
  title={#3},
  colframe=#1,
  colback=#2,
  colbacktitle=#1,
  coltitle=white,
  fonttitle=\bfseries\large,
  fontupper=\rmfamily,
  arc=1.5mm,
  boxrule=1.2pt,
  top=3mm, bottom=3mm, left=3mm, right=3mm,
  toptitle=0.5mm, bottomtitle=0.5mm,
  before upper={\setlength{\parindent}{1.5em}}
}

\newcounter{idx}

\icmltitlerunning{\Benchmark: An Evaluation Benchmark for Tool-Augmented Agentic Reasoning on Scientific Literature}

\begin{document}

\twocolumn[
  \icmltitle{PaperArena: An Evaluation Benchmark for Tool-Augmented \\ Agentic Reasoning on Scientific Literature}



  \icmlsetsymbol{equal}{*}

  \begin{icmlauthorlist}
    \icmlauthor{Daoyu Wang}{ustc}
    \icmlauthor{Mingyue Cheng$^*$}{ustc}
    \icmlauthor{Shuo Yu}{ustc}
    \icmlauthor{Zirui Liu}{ustc}
    \icmlauthor{Ze Guo}{ustc}
    \icmlauthor{Xin Li}{ustc,iflytex}
    \icmlauthor{Qi Liu}{ustc}
  \end{icmlauthorlist}

  \icmlaffiliation{ustc}{State Key Laboratory of Cognitive Intelligence, University of Science and Technology of China, Hefei, China}
  \icmlaffiliation{iflytex}{Artificial Intelligence Research Institute, iFLYTEK Co., Ltd, Hefei, China}

  \icmlcorrespondingauthor{Mingyue Cheng}{mycheng@ustc.edu.cn}

  \icmlkeywords{Machine Learning, ICML}

  \vskip 0.3in
]



\printAffiliationsAndNotice{}  

\begin{abstract}
Understanding and reasoning on the large-scale scientific literature is a crucial touchstone for large language model (LLM) based agents. However, existing works are mainly restricted to tool-free tasks within single papers, largely due to the lack of a benchmark that evaluates cross-paper reasoning and multi-tool orchestration in authentic research scenarios. In this work, we propose \Benchmark, a benchmark to evaluate LLM-based agents on questions that require integrating information across multiple papers with the assistance of external tools. Given a research question, agents should formulate a reasoning plan, interact with multiple papers, and invoke appropriate tools to produce a well-grounded answer. To support standardized evaluation, we provide a platform for agent execution, offering a modular tool environment including multimodal parsing, context retrieval, and programmatic computation. Experiments reveal that even the leading LLM powering a well-established agentic workflow achieves merely 38.78\% average accuracy, while on the hard subset, accuracy drops to only 18.47\%. We also analyze reasoning traces and diagnose agent behavior, providing the community with insights to develop and evaluate more capable scientific agents.
Our code and data are available at 
\url{https://github.com/Melmaphother/PaperArena}.
\end{abstract}

\section{Introduction}

\begin{figure}[t]
    \centering
    \includegraphics[width=1\linewidth]{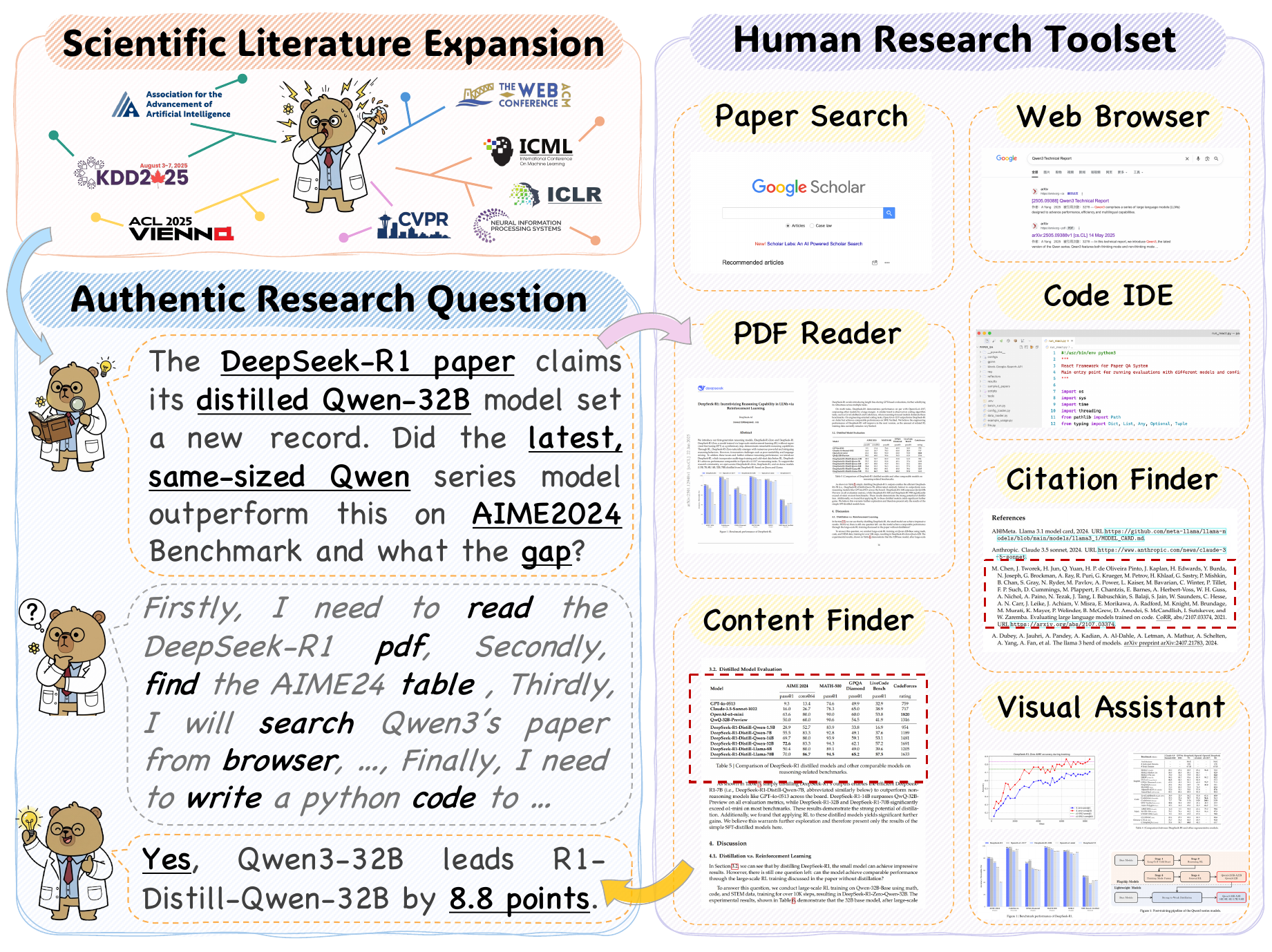}
    \caption{Facing the expanding scientific literature, researchers must perform rigorous reasoning and utilize diverse tools to address complex questions in authentic research scenarios.}
    \label{fig:fig_1}
\end{figure}

With the large-scale scientific literature continually expanding, researchers are increasingly challenged by information overload~\cite{houssard2024gerontocratization,bollacker2002discovering}. As shown in Figure~\ref{fig:fig_1}, addressing an authentic research question is a cognitively demanding task that requires researchers to formulate plans, leverage tools, and interact with heterogeneous content across multiple papers~\cite{kononova2021opportunities,katz2024knowledge}. In response to this challenge, a growing trend is to use LLM-based agents to automate these complex workflows~\cite{gao2025tooluniverse,zhang2025chain}.

Despite this potential, progress in LLM-based agents for scientific research remains limited. A primary obstacle is the lack of a benchmark designed to evaluate agents in addressing complex authentic research questions. As shown in Table~\ref{tab:tab_1}, existing benchmarks on scientific literature are often limited to localized tasks such as explaining a concept from a single paragraph or extracting a specific value from a chart~\cite{jin2019pubmedqa, pramanick2024spiqa}. However, these overly simplified tasks are often solvable by the inherent capabilities of modern LLMs alone, failing to provide a sufficiently challenging testbed for evaluating the tool-augmented reasoning capabilities of advanced agents~\cite{yao2023react,cheng2025survey,li2025llms}.

This critical gap motivates our pioneering effort to construct an evaluation benchmark specifically for assessing LLM-based agents reasoning on scientific literature. This endeavor is challenging for several reasons. First, the construction process is non-trivial and labor-intensive, requiring cross-paper synthesis, fine-grained citation matching, and structural multimodal parsing~\cite{mialon2023gaia,siegel2024core}. Second, evaluating agents poses unique challenges from traditional LLM assessment, necessitating a dedicated tool environment and modular orchestration of planning, tool interaction and long-context management~\cite{yue2025survey,starace2025paperbench}. Finally, meaningful evaluation must extend from final answer accuracy to examination of intermediate reasoning traces and diagnostics of agent behavior~\cite{wang2022neuralcd}.

With these challenges in mind, we propose \Benchmark\ and its accompanying \Platform. The construction of \Benchmark\ begins with a vast corpus comprising tens of thousands of papers from premier open-access AI conferences. From this corpus, we sample a representative subset and employ LLMs to synthesize initial question drafts. To ensure high complexity, a team of expert annotators filter and rewrite these drafts into authentic formats and label their answers, tool chains, difficulty, and question types. This rigorous pipeline yields a final set of 784 high-quality questions. Concurrently, we provide \Platform~to support the evaluation of both single- and multi-agent systems. The platform offers a modular tool environment including multimodal parsing, context retrieval, and code execution, while also managing the complete agent lifecycle.

We conduct extensive experiments on various leading LLM-based agents. 
Results reveal a stark performance gap: the advanced LLM, Gemini 2.5 Pro~\cite{comanici2025gemini}, powering a well-established multi-agent system achieves an average accuracy of only 38.78\% and drops to 18.47\% on the hard subset, a performance far below the 83.5\% achieved by a human expert.
Beyond overall accuracy, we observe inefficient tool usage and fragile plan execution in intermediate reasoning traces. Further diagnosis via item response theory (IRT)~\cite{liu2024computerized} reveals limitations in latent ability of current LLM-based agents and their distinct failure patterns. We invite the community to adopt \Benchmark\ and \Platform\ to develop and evaluate more capable agents for scientific discovery and to accelerate progress in agent-driven research.


In summary, our main contributions are as follows: \begin{itemize} 
    \item We introduce \Benchmark, a challenging benchmark for tool-augmented agentic reasoning on scientific literature, integrated with \Platform\ to enable standardized evaluation across diverse workflows.
    \item We conduct extensive evaluations of leading LLM-based agents. The results reveal a significant performance gap between current agents and human experts.
    \item By analyzing intermediate reasoning traces and diagnosing of agent behavior, we identify distinct agent strengths and specific failure patterns.
\end{itemize}

\begin{table}[!t]
    \centering
    \caption{Comparing \Benchmark\ with existing scientific QA benchmarks across multi-step reasoning, multimodal understanding, cross-paper integration, and database interfacing.}
    \label{tab:tab_1}
    \small
    \resizebox{1\linewidth}{!}{%
        \begin{tabular}{l  cccc}
            \toprule
            \textbf{Benchmark} & 
            \textbf{\begin{tabular}[c]{@{}c@{}}Multi-Step \\ Reasoning\end{tabular}} & 
            \textbf{\begin{tabular}[c]{@{}c@{}}Multimodal \\ Understanding\end{tabular}} & 
            \textbf{\begin{tabular}[c]{@{}c@{}}Cross-paper \\ Integration\end{tabular}} & 
            \textbf{\begin{tabular}[c]{@{}c@{}}Database \\ Interfacing\end{tabular}} \\
            \midrule
            PubMedQA~\cite{jin2019pubmedqa} & \xmark & \xmark & \xmark & \xmark \\
            CharXiv~\cite{wang2024charxiv} & \cmark & \cmark & \xmark & \xmark \\
            emrQA~\cite{pampari2018emrqa} & \xmark & \xmark & \cmark & \xmark \\
            QASA~\cite{lee2023qasa} & \xmark & \xmark & \xmark & \xmark \\
            ArgSciChat~\cite{ruggeri2023dialogue} & \cmark & \xmark & \xmark & \xmark \\
            QASPER~\cite{dasigi2021qasper} & \xmark & \xmark & \xmark & \xmark \\
            SPIQA~\cite{pramanick2024spiqa} & \cmark & \cmark & \xmark & \xmark \\
            \midrule
            \textbf{\Benchmark} & \cmark & \cmark & \cmark & \cmark \\
            \bottomrule
        \end{tabular}
    }
\end{table}

\section{Related Work}

\subsection{Benchmarks on Scientific Literature}
Benchmarks for scientific literature understanding and reasoning grow rapidly in recent years \cite{hendrycks2020measuring, saikh2022scienceqa}. Early works focus on localized tasks, such as claim verification and phrase extraction \cite{wadden2020fact, augenstein2017semeval}. With the rise of LLMs, research attention shifts toward knowledge-intensive scientific data, particularly structured academic charts \cite{zhou2025benchmarking}. Driven by the potential of LLM-based agents, benchmarks now target from-scratch scientific reproduction \cite{hua2025researchcodebench, starace2025paperbench}. Subsequent works expand into specialized fields like biomedicine and chemistry, prioritizing domain-specific professional challenges \cite{jin2019pubmedqa, krithara2023bioasq, guo2023can}. To better simulate authentic research, modern evaluation incorporates autonomous knowledge discovery, peer-review critique, and expert-level exams \cite{majumder2024discoverybench, baumgartner2025peerqa, phan2025humanity}.

\subsection{Tool-Augmented Agents}
The evolution of LLM reasoning capabilities enable complex agent workflows \cite{wei2022chain, achiam2023gpt, li2025llms}. Early efforts expand reasoning boundaries through external web-based retrieval and synthesis \cite{nakano2021webgpt,ahn2022can}. As reasoning and action capabilities converge, agents evolve further through planning, tool orchestration, memory management, and iterative self-reflection. \cite{yao2023react,shinn2023reflexion,huang2024understanding,li2025memos,qiu2025alita}. Furthermore, multi-agent systems push the frontiers of collaborative problem-solving \cite{hong2023metagpt, wu2024autogen, zhang2025metaagent, wang2025cooperative}. 
Regarding agent evaluation, current benchmarks assess agents on literature retrieval or tool-based biological research \cite{lala2023paperqa,skarlinski2024language,jin2025stella}. However, most existing works remain restricted to tool-free tasks within single papers, lacking effective benchmarks for tool-centric, complex literature understanding and reasoning. To bridge this gap, we propose the \Benchmark\ benchmark and its accompanying evaluation platform \Platform.

\section{\Benchmark}
In this section, we first establish a clear task formulation, followed by a detailed construction of \Benchmark. Furthermore, we provide an in-depth analysis of key features that distinguish \Benchmark\ from existing benchmarks.

\subsection{Task Formulation}
\Benchmark\ tasks an agent with an authentic research question $q$ and a source paper $P$ in \textit{pdf} format. The agent must formulate a plan and iteratively leverage a predefined external toolset $\mathcal{T}$. The reasoning process entails interaction with diverse content particularly when $q$ involves a set of associated papers $\mathcal{P}_{\text{assoc}}$. The agent processes the observation from the tool execution to determine whether to invoke a subsequent tool or yield the final answer to $q$.

\subsection{Benchmark Construction}
\label{sec:3.2}
\begin{figure}[t]
    \centering
    \includegraphics[width=1\linewidth]{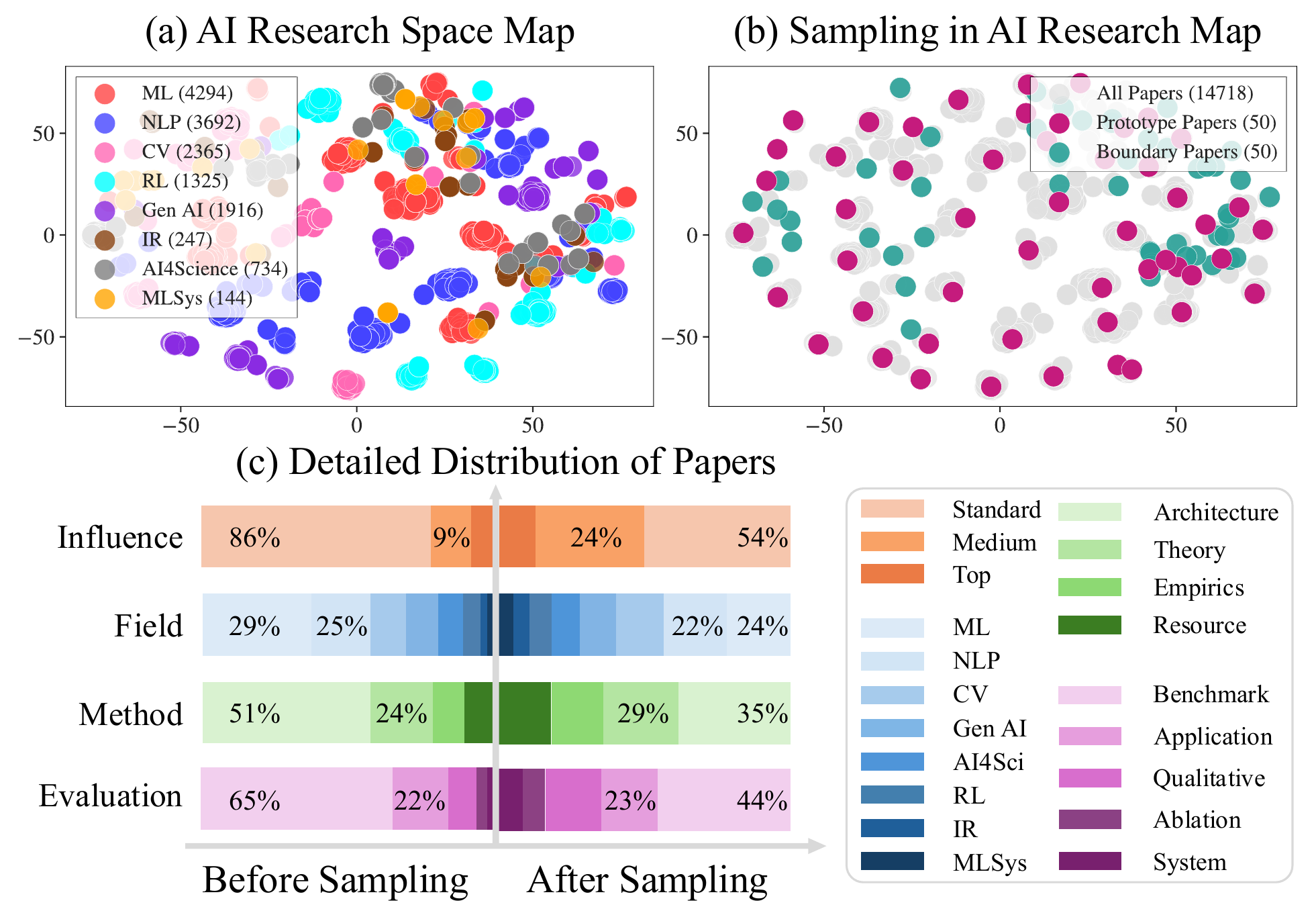}
    \caption{Our sampling strategy helps build a balanced paper subset, as shown by the distribution changes across different research fields before and after the sampling process.}
    \label{fig:fig_2}
\end{figure}

\paragraph{Paper Corpus Preparation.}
We construct \Benchmark\ on a curated corpus of 14,435 open-access latest AI papers from OpenReview \cite{openreview} and Open Access \cite{openaccess}. Each paper is labeled with a one-hot vector encoding its influence, research field, method category and evaluation type. A t-SNE visualization \cite{maaten2008visualizing} of these vectors (Figure~\ref{fig:fig_2}(a)) reveals strong domain-level clustering, indicating topic redundancy and the risk of evaluation bias. To address this, we adopt a hierarchical sampling strategy. We first apply K-Medoids \cite{martins2024k} to identify 50 prototype papers that reflect dominant research trends. Meanwhile, to ensure broader coverage, we further apply farthest point sampling (FPS) to select 50 boundary papers from underrepresented regions. As shown in Figure~\ref{fig:fig_2}(b) and (c), the resulting 100-paper subset balances representativeness and diversity.

\paragraph{QA Generation from Papers.}
We first define the external toolset $\mathcal{T}$ to align the generated questions with solvable operational boundaries, encompassing web searcher, cross-ref searcher, pdf parser, context retriever, figure/table analyzers, code executor, and database querier. To preserve critical structural information such as section headers and citations, we convert each paper's \textit{pdf} into \textit{markdown} format using MinerU \cite{wang2024mineru}. We then prompt three distinct advanced LLMs with these structured segments and the specific definitions of $\mathcal{T}$ to independently generate candidate questions, yielding a total of 2,875 initial drafts. These LLM-generated drafts serve solely as preliminary seeds. A team of Ph.D. experts rigorously processes these drafts by filtering out trivial entries, balancing tool usage frequencies, removing obvious indicators to ensure sufficient difficulty and ultimately rewriting the drafts into authentic research question format. Finally, each question is carefully annotated with the ground-truth answer, the minimal viable tool chain, the difficulty level (easy, medium, hard), and the question type (multiple choice, concise answer, open answer). Further elaboration of this human-in-the-loop process can be found in Appendix \ref{sec:a}.

\begin{figure*}[t]
    \centering
    \includegraphics[width=1\linewidth]{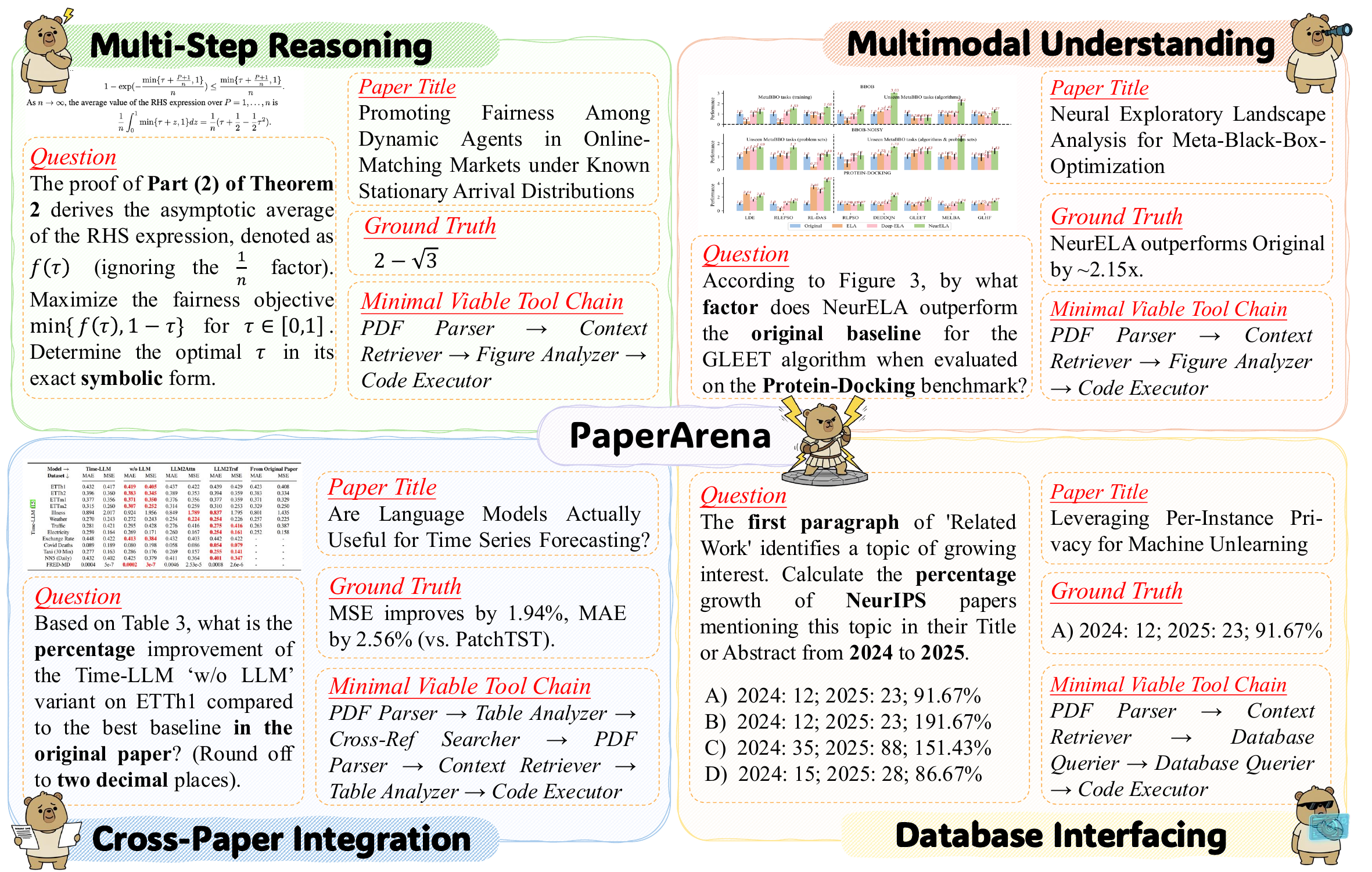}
    \caption{Cases of the four key features in \Benchmark\ comprising multi-step reasoning, multimodal understanding, cross-paper integration, and database interfacing. Each quadrant presents a sample question with the associated paper title, ground truth, and the minimal viable tool chain. These samples illustrate the necessity of tool-augmented reasoning agent on \Benchmark.}
    \label{fig:fig_3}
\end{figure*}

\subsection{Key Features of \Benchmark}

Unlike existing benchmarks, \Benchmark\ requires multi-tool and cross-paper reasoning. As shown in Figure~\ref{fig:fig_3}, we describe the four key features of \Benchmark:

\paragraph{Multi-Step Reasoning.} 
This category involves complex theoretical problems that require sequential deduction. The illustrated example demands deriving a symbolic solution for a mathematical theorem to maximize an objective function. Such tasks necessitate that the agent execute multi-step reasoning and utilize code execution tools to ensure mathematical precision when handling abstract logic and formal proofs within scientific literature.

\paragraph{Multimodal Understanding.}
This feature targets the interpretation of scientific charts alongside textual context. In the shown case the agent must quantify performance gaps by extracting specific values from a bar chart. This demonstrates the need for the agent to align visual information with specific queries and perform precise quantitative analysis on complex graphical representations of experimental data.

\paragraph{Cross-Paper Integration.}
This dimension addresses the challenge of synthesizing fragmented information from multiple papers. The provided sample requires calculating percentage improvements by comparing local results against baselines in an external citation. This scenario highlights the necessity for the agent to navigate between papers and integrate cross-reference context to verify claims and perform comparative analysis across distinct sources.

\paragraph{Database Interfacing.}
This aspect focuses on retrieving and processing data from external repositories. The example task involves determining the growth rate of a specific topic by analyzing metadata from conference proceedings. It illustrates the requirement for the agent to interface with databases via executable queries to extract statistical information from large structured database.

\begin{table*}[th!]
    \centering
    \setlength{\tabcolsep}{4pt}
    \caption{Comprehensive evaluation of nine leading LLMs in single- and multi-agent workflows on \Benchmark\ against Ph.D. experts on \Benchmark-Human. Performance is stratified by question type and difficulty level. The best accuracy (\%) is in bold. We also report the average number of reasoning steps and reasoning efficiency (\%).}
    \label{tab:tab_2}
    \resizebox{\textwidth}{!}{
        \fontsize{10}{12}\selectfont
        \renewcommand{\arraystretch}{1.15}
        \begin{tabular}{l cc  ccc ccc ccc  c cc}
            \toprule
            \multirow{2}{*}{\textbf{Base LLM}} & \multirow{2}{*}{\textbf{Thinking}} & \multirow{2}{*}{\textbf{Open Src.}} & \multicolumn{3}{c}{\textbf{Multiple Choice}} & \multicolumn{3}{c}{\textbf{Concise Answer}} & \multicolumn{3}{c}{\textbf{Open Answer}} & \multicolumn{3}{c}{\textbf{Average}} \\
            \cmidrule(lr){4-6}
            \cmidrule(lr){7-9}
            \cmidrule(lr){10-12}
            \cmidrule(lr){13-13}        
            \cmidrule(lr){14-14}
            \cmidrule(lr){15-15}
            &  &  &  \hspace{0.6em} Easy \hspace{0.6em}  & Medium &  \hspace{0.6em} Hard \hspace{0.6em} & \hspace{0.6em} Easy \hspace{0.6em} & Medium & \hspace{0.6em} Hard \hspace{0.6em} & \hspace{0.6em} Easy \hspace{0.6em} & Medium & \hspace{0.6em} Hard \hspace{0.6em} & \hspace{0.6em} Acc. \hspace{0.6em} &  \hspace{0.2em} Steps \hspace{0.2em} &  \hspace{0.6em} Eff.  \hspace{0.6em} \\
            \midrule
            \multicolumn{15}{c}{ \textit{\Platform~(Single-Agent)}} \\
            \midrule
            Gemini 2.5 Pro & \cmark & \xmark & 57.89 & 40.98 & \textbf{22.22} & \textbf{58.49} & \textbf{40.00} & \textbf{21.95} & \textbf{56.52} & \textbf{36.07} & 13.57 & \textbf{36.10} & 9.81 & 41.20 \\
            OpenAI o4-mini-high & \cmark & \xmark & 57.89 & \textbf{44.26} & \textbf{22.22} & 55.66 & 37.30 & 14.63 & 52.17 & 33.61 & 12.86 & 33.93 & 10.63 & 35.12 \\
            Claude Sonnet 4 & \cmark & \xmark & \textbf{60.53} & 37.70 & 18.52 & 53.77 & 38.92 & 19.51 & 52.17 & 31.97 & \textbf{15.00} & 34.18 & 9.55 & 37.34 \\
            GPT-4.1 & \xmark & \xmark & 55.26 & 34.43 & 14.81 & 49.06 & 35.68 & 17.07 & 52.17 & 28.69 & 10.71 & 30.61 & 8.27 & 35.26 \\
            Claude 3.5 Sonnet & \xmark & \xmark & 47.37 & 28.23 & 11.11 & 48.11 & 25.95 & 10.98 & 43.48 & 22.95 & 7.14 & 24.62 & 7.89 & 21.98 \\
            Qwen3-235B-Thinking & \cmark & \cmark & 52.63 & 32.79 & 14.81 & 53.77 & 33.51 & 15.85 & 47.83 & 27.87 & 10.00 & 29.97 & 10.20 & 31.21 \\
            GLM-4.5 & \cmark & \cmark & 47.37 & 29.51 & 11.11 & 49.06 & 30.27 & 13.41 & 43.48 & 27.05 & 8.57 & 27.17 & 11.82 & 25.13 \\
            Qwen3-235B-Instruct & \xmark & \cmark & 44.74 & 24.59 & 11.11 & 46.23 & 24.32 & 10.98 & 39.13 & 22.13 & 7.14 & 23.47 & 7.24 & 36.85 \\
            Kimi-K2-Instruct & \xmark & \cmark & 44.74 & 22.95 & 11.11 & 45.28 & 22.70 & 9.76 & 39.13 & 20.49 & 6.43 & 22.32 & 12.86 & 23.57 \\
            \midrule
            \multicolumn{15}{c}{ \textit{\Platform~(Multi-Agent)}} \\
            \midrule
            Gemini 2.5 Pro & \cmark & \xmark & 60.53 & \textbf{42.62} & \textbf{33.33} & \textbf{63.21} & \textbf{42.70} & 21.95 & \textbf{65.22} & \textbf{39.34} & 13.57 & \textbf{38.78} & 8.58 & 45.39 \\
            OpenAI o4-mini-high & \cmark & \xmark & 60.53 & 41.00 & 25.93 & 62.26 & 41.08 & 21.95 & 56.52 & 36.89 & \textbf{14.29} & 37.37 & 8.30 & 43.41 \\
            Claude Sonnet 4 & \cmark & \xmark & \textbf{68.42} & 39.34 & 18.52 & 61.32 & 40.00 & \textbf{23.17} & 56.52 & 36.07 & 12.86 & 36.73 & 8.92 & 39.13 \\
            GPT-4.1 & \xmark & \xmark & 57.89 & 37.70 & 18.52 & 59.43 & 38.38 & 19.51 & 52.17 & 34.43 & 12.14 & 34.57 & 7.94 & 34.35 \\
            Claude 3.5 Sonnet & \xmark & \xmark & 50.00 & 31.15 & 14.81 & 52.83 & 32.43 & 14.63 & 47.83 & 29.51 & 9.29 & 29.34 & 8.16 & 23.67 \\
            Qwen3-235B-Thinking & \cmark & \cmark & 57.89 & 37.70 & 18.52 & 59.43 & 37.84 & 19.51 & 52.17 & 33.61 & 12.14 & 34.31 & 10.48 & 33.79 \\
            GLM-4.5 & \cmark & \cmark & 52.63 & 32.79 & 14.81 & 54.72 & 34.05 & 15.85 & 47.83 & 30.33 & 10.71 & 30.74 & 10.90 & 27.01 \\
            Qwen3-235B-Instruct & \xmark & \cmark & 50.00 & 31.15 & 14.81 & 51.89 & 31.35 & 14.63 & 47.83 & 28.69 & 9.29 & 28.83 & 7.82 & 38.13 \\
            Kimi-K2-Instruct & \xmark & \cmark & 50.00 & 27.87 & 11.11 & 50.00 & 28.11 & 12.20 & 43.48 & 25.41 & 7.86 & 26.28 & 11.94 & 29.45 \\
            \midrule
            \multicolumn{15}{c}{\textit{Human Baseline (\Benchmark-Human)}} \\
            \midrule
            Ph.D. Experts & - & - & \cellcolor{gray!15}87.50 & \cellcolor{gray!15}84.21 & \cellcolor{gray!15}71.43 & \cellcolor{gray!15}91.67 & \cellcolor{gray!15}85.42 & \cellcolor{gray!15}68.18 & \cellcolor{gray!15}87.50 & \cellcolor{gray!15}86.67 & \cellcolor{gray!15}82.35 & \cellcolor{gray!15}83.50 & \cellcolor{gray!15}6.52 & \cellcolor{gray!15}75.48 \\
            \bottomrule
        \end{tabular}
    }
\end{table*}

\section{\Platform}
Tailored for the unique requirements of agent evaluation, we present \Platform. This modular platform enables standardized evaluation on \Benchmark, incorporating agent-centric architecture and specialized evaluation settings.

\subsection{Platform Architecture}

\paragraph{Agentic Workflow.} 
\Platform\ supports both single- and multi-agent workflows. In the single-agent setting, we adopt the ReAct-style workflow \cite{yao2023react}, where LLMs reason and act in alternation. In the multi-agent setting, we employ a centralized design with a manager agent responsible for high-level planning and delegating subtasks to worker agents within independent context spaces. Regardless of the setting, \Platform\ manages the entire reasoning lifecycle of the agents.

\paragraph{Tool Environment.}
\Platform\ integrates the full suite of tools defined in Section \ref{sec:3.2}, allowing agents to interact with multiple papers, databases, and external resources. For example, our implementation of the database querying tool is backed by a structured dataset covering metadata from hundreds of major AI conferences in recent years. This includes paper titles, abstracts, and links. Agents can perform operations such as retrieving all papers on a topic from a specific conference and year, or extracting relevant entries based on keyword queries. Additionally, we incorporate robust error handling to minimize tool execution failures, ensuring reliable performance during long tool chains.

\subsection{Evaluation Setup}
\paragraph{Agent Baselines.}
We evaluate representative nine leading LLMs in both single-agent and multi-agent workflows: Gemini 2.5 Pro~\cite{comanici2025gemini}, OpenAI o4-mini-high~\cite{OpenAI2025o4mini}, Claude Sonnet 4~\cite{Anthropic2025Claude4}, GPT-4.1~\cite{OpenAI2025gpt41}, Claude 3.5 Sonnet~\cite{Anthropic2025Claude35}, Qwen3-235B-Thinking~\cite{yang2025qwen3}, GLM-4.5~\cite{zeng2025glm45}, Qwen3-235B-Instruct~\cite{yang2025qwen3}, and Kimi-K2-Instruct~\cite{teamk2025kimi}. We construct a 200-question subset from \Benchmark~(\Benchmark-Human) and recruit three Ph.D. candidates, distinct from the annotators, to complete it using tools like GPT-4o \cite{hurst2024gpt4o} and google search tools.

\paragraph{Evaluation Metrics.}
We adopt multi-dimensional metrics to evaluate agent performance. 
Firstly, regarding outcome accuracy, all answers are evaluated using the LLM-as-a-Judge~\cite{liu2025elo}, where GPT-4o assigns binary labels. We prioritize validity over formatting, allowing for flexible output styles instead of strict exact matching~\cite{wang2025can}. 
Secondly, to analyze the reasoning process, we measure: (1) reasoning steps, the mean length of executed tool chains; and (2) efficiency, the normalized overlap between the tool chain executed by the agent and the minimal viable chain.
Crucially, going beyond aggregate statistics, we employ item response theory (IRT)~\cite{liu2024computerized} to diagnose agent behavior. Specifically, we utilize a response matrix that records the binary correctness for each agent-question pair. By applying markov chain monte carlo (MCMC) sampling, we infer the item difficulty parameters \(\beta\) and latent agent ability \(\theta\) based on these responses.  Finally, to mitigate potential biases in LLM-as-a-judge, we manually verify all human answers and LLM responses to the questions within the \Benchmark-Human. Formal definitions for the metrics are available in Appendix~\ref{sec:b}.

\begin{figure}
    \centering
    \includegraphics[width=1\linewidth]{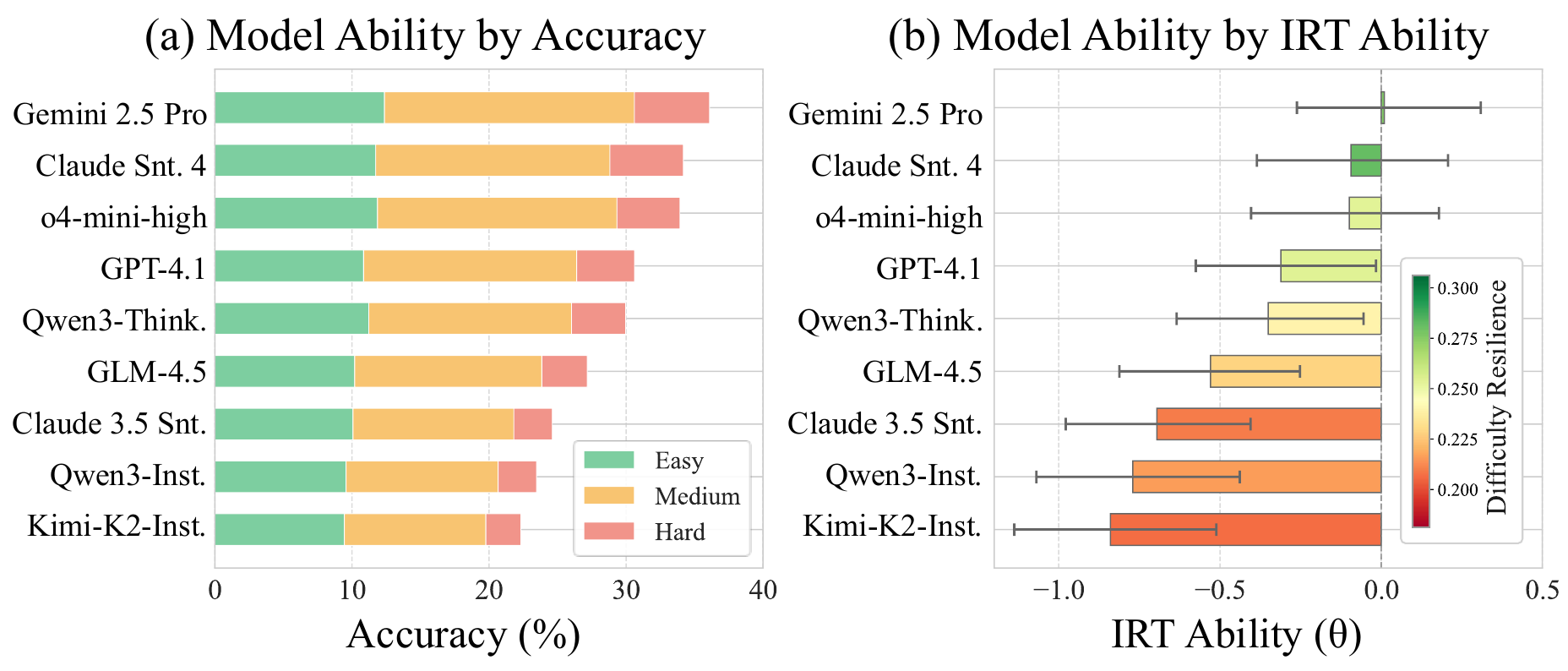}
    \caption{Comparison of model ability using accuracy versus IRT estimation based on solely difficulty level, showing a positive correlation while visualizing difficulty resilience of different LLMs.}
    \label{fig:fig_4}
\end{figure}

\section{Experiments}

In this section, we evaluate the overall performance of the agents and present three key observations regarding their reasoning processes. Finally, we conduct an in-depth diagnosis of model capabilities and distinctive failure patterns.

\subsection{Main Results}

Table~\ref{tab:tab_2} summarizes the performance of nine leading LLMs under both single- and multi-agent workflows. Gemini~2.5~Pro achieves the best overall results in both settings, yet a substantial performance gap persists between all LLM-based agents and the Ph.D. expert baseline.  Performance consistently degrades in harder tasks and open-ended formats due to increased reasoning demands. Moreover, proprietary LLMs generally outperform open-source counterparts across most metrics. LLMs with deep thinking ability achieve higher accuracy than non-thinking ones, as the reasoning process enables implicit task decomposition and intermediate tool planning. While multi-agent workflows improve accuracy and efficiency by enabling collaborative reasoning and labor division, most agents still lack planning efficiency, often invoking tools far more than necessary.
Additionally, Figure~\ref{fig:fig_4} illustrates a positive correlation between accuracy and IRT estimation. Nevertheless, taking Claude 3.5 Sonnet as an example, its varied difficulty resilience shows that high accuracy does not strictly guarantee stability on complex tasks, underscoring the importance of implicit capability assessment.

\begin{figure}[t]
    \centering
    \includegraphics[width=1\linewidth]{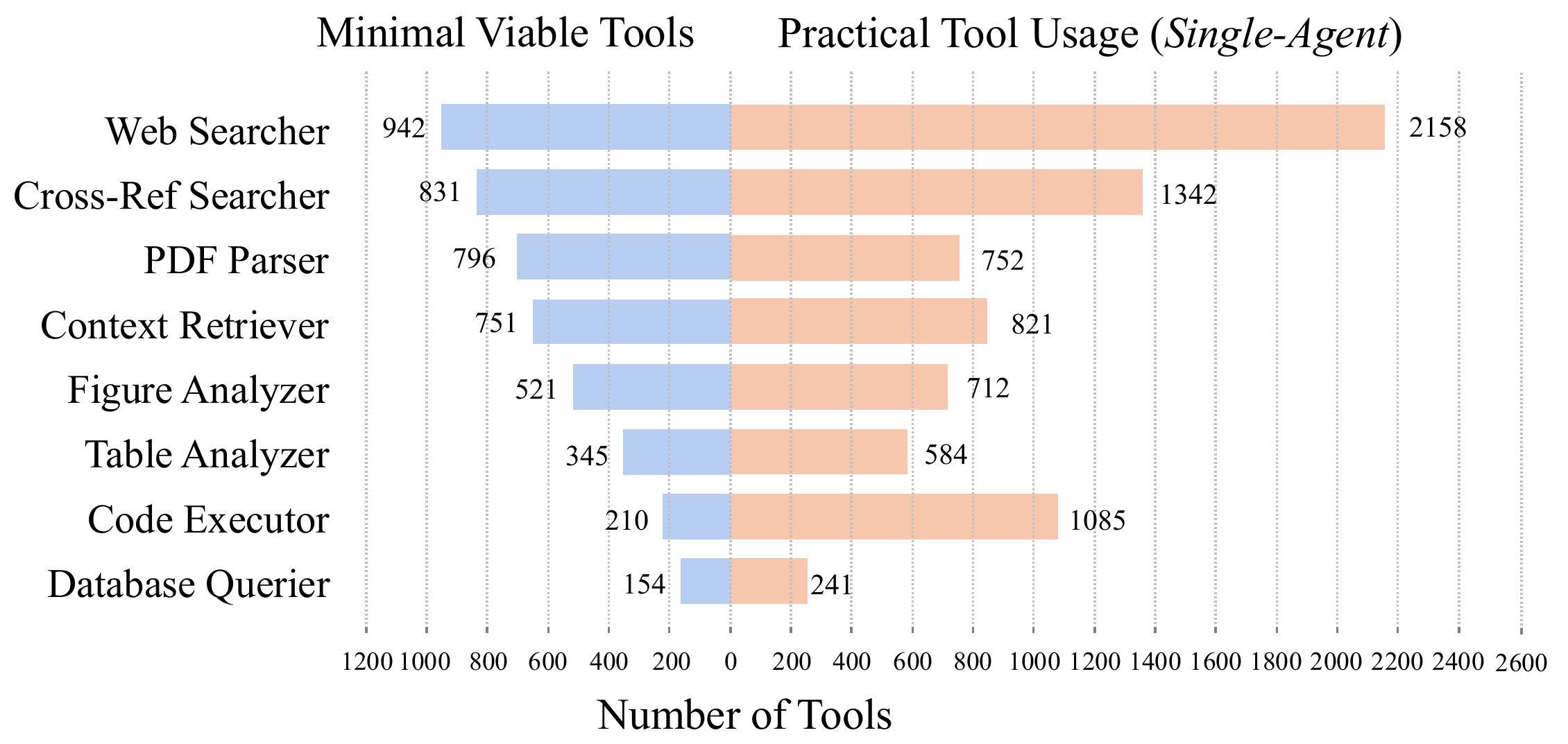}
    \caption{Comparison of minimal viable and practical tool usage by Gemini 2.5 Pro in the single-agent workflow, demonstrating a substantial tool usage imbalance between the two.}
    \label{fig:fig_5}
\end{figure}

\subsection{Observation of Reasoning Traces}

Beyond measuring accuracy, we monitor reasoning traces and observe three significant phenomena that provide insights for the future progress of agents.

\paragraph{Tool Usage Imbalance.} Agent tool usage is strongly imbalanced, showing a clear preference for general-purpose tools like the web searcher and code executor over more specialized ones. Figure~\ref{fig:fig_5}, which details the gap between tools in the practical reasoning trace and the minimal viable tool chain, clearly illustrates this phenomenon. This biased preference is a primary driver of the significant efficiency gap. For instance, practical use of the code executor is over five times the minimal viable requirement, suggesting a default to a brute-force, trial-and-error coding approach. Similarly, practical usage of the web searcher is more than double the minimal viable baseline, indicating that the imbalanced strategy results in inefficient, exploratory searches rather than precise queries. In contrast, for deterministic tasks requiring a specific, non-preferred tool like the pdf parser, the practical and minimal viable calls are nearly identical. This suggests that the observed inefficiency is not a universal failure of capabilities, but rather a direct consequence of the imbalanced tool selection strategy.

\begin{figure}[t]
    \centering
    \includegraphics[width=1\linewidth]{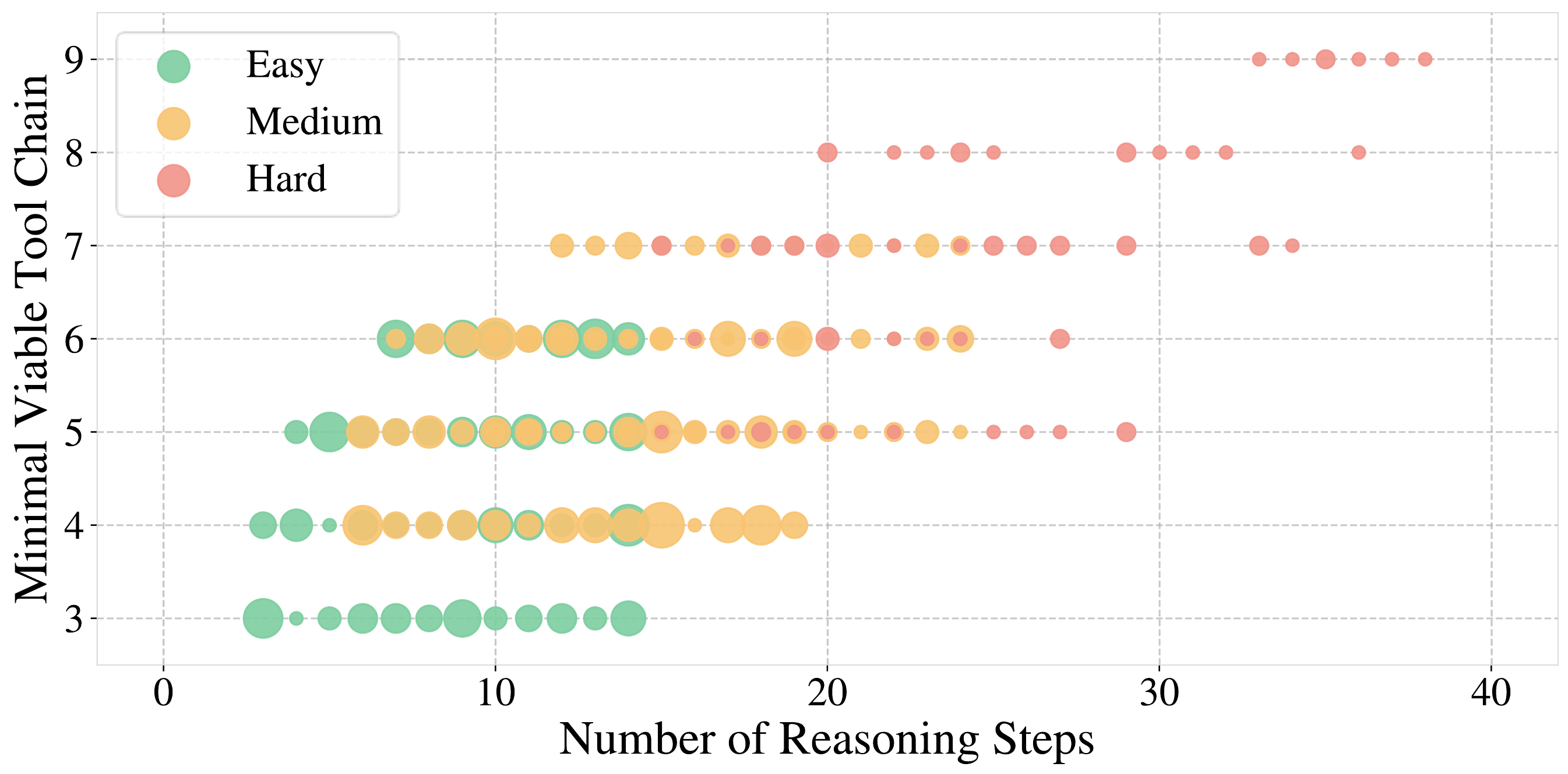}
    \caption{Positive correlation between the minimal viable tool chain and reasoning steps across three difficulty levels, showing that harder tasks consistently demand more extensive reasoning.}
    \label{fig:fig_6}
\end{figure}

\begin{figure}[t]
    \centering
    \includegraphics[width=1\linewidth]{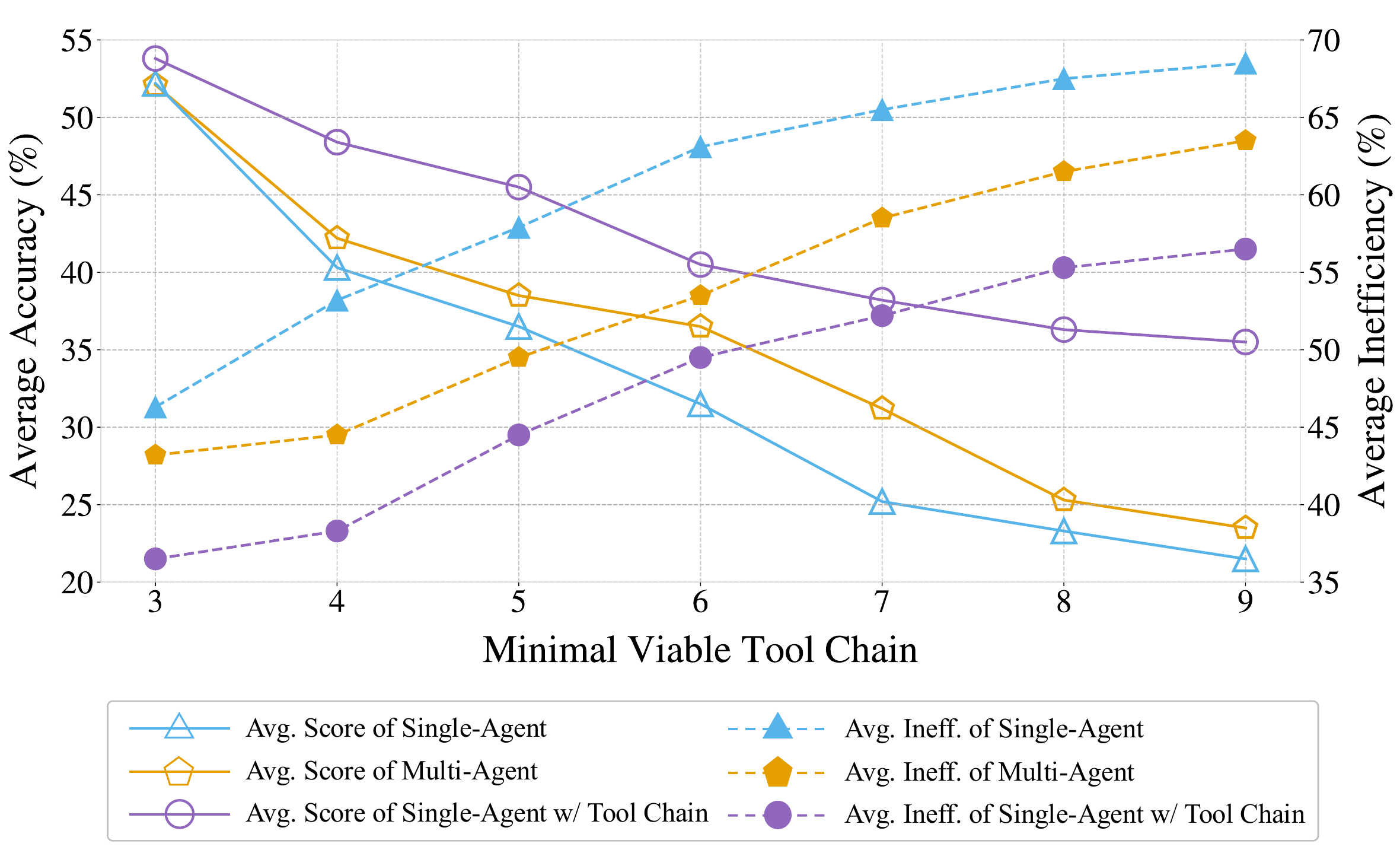}
    \caption{Efficiency analysis across tool chain lengths. Solid lines denote Score; dashed lines show average inefficiency ($1 - \text{efficiency}$), quantifying the degree of ineffective tool usage.}
    \label{fig:fig_7}
\end{figure}

\paragraph{Reasoning Bottleneck.} Agents tend to employ more reasoning steps for more difficult problems, yet the marginal utility of this extended reasoning diminishes, failing to avert a decline in accuracy. Figure~\ref{fig:fig_6} illustrates a positive correlation between minimal viable tool chain length and reasoning steps. While harder tasks demand longer reasoning, this increased effort yields disproportionately low performance gains. As illustrated in Figure~\ref{fig:fig_7}, while the accuracy of both single- and multi-agent workflows declines with increasing task complexity, multi-agent workflows consistently maintain a performance advantage over single-agent baselines in complex, multi-tool scenarios. This enhanced performance highlights the structural advantage of the centralized design, where the manager agent delegates concrete execution to worker agents. By decoupling high-level planning from specific tool invocations, the manager maintains a clean reasoning context free from execution noise, thereby ensuring more effective and stable planning.

\begin{figure}
    \centering
    \includegraphics[width=1\linewidth]{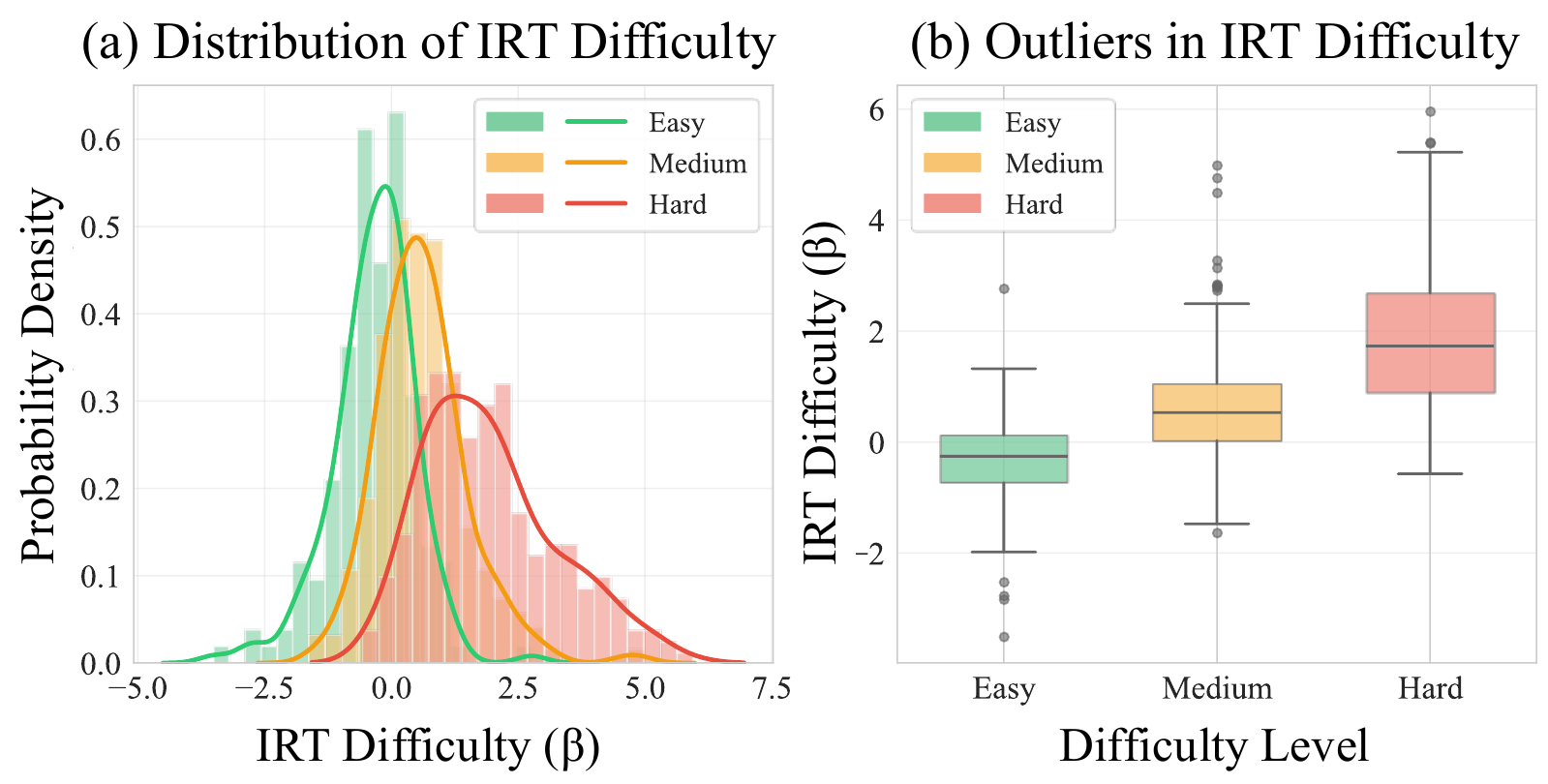}
    \caption{The distribution of IRT difficulty demonstrates the validity of assigned difficulty level while outlier points indicate significant discrepancies between agent ability and human expectations.}
    \label{fig:fig_8}
\end{figure}

\paragraph{Execution Inefficiency.} The single agent, when explicitly prompted with the globally minimal viable tool chain, outperforms multi-agent workflows. This observation implies that endogenous planning in current LLMs is susceptible to local optima and warrants further investigation. We conduct this experiment by prompting the single agent with the minimal viable tool chain before it starts reasoning. As shown in Figure~\ref{fig:fig_7}, the minimal viable tool chain agent establishes a near-perfect planning ceiling, leaving a consistent gap from the single- and multi-agent workflow. The efficiency analysis further illuminates this: while multi-agent planning (yellow line) is more effective than the single-agent baseline (blue line), it exhibits substantial ineffectiveness compared to the ceiling established by the minimal viable plan (purple line). Crucially, the agent exhibits increasing inefficiency with more complex tasks, even when utilizing a minimal viable tool chain. This identifies flawed tool invocation as a second distinct failure mode, particularly when agents fail to execute tools correctly despite having a near-perfect plan, such as by providing incorrect parameters.

\subsection{Diagnostics for Agent Behavior}

Beyond quantitative metrics, we diagnose problem difficulty and agent capabilities. This analysis provides insights into agent behavior and reveals unique failure modes.

\paragraph{IRT-Based Agent Ability.} To quantitatively evaluate the complexity of tasks within \Benchmark, we estimate the IRT difficulty parameter $\beta$ for each question. As illustrated in Figure~\ref{fig:fig_8} (a), the probability density distributions for easy, medium, and hard difficulty levels are well-separated with clear peaks, which confirms that the manual difficulty annotations align well with empirical agent performance across all tested questions. The sequential shift of these bell-shaped curves toward higher $\beta$ values validates the structural integrity of the difficulty stratification. However, Figure~\ref{fig:fig_8} (b) reveals the presence of a certain number of upper outliers, particularly within the medium difficulty level. These high-value outliers show specific cases where agents fail unexpectedly on conceptually simple tasks, suggesting a clear gap between current agent capabilities and human expectations. These results indicate that while the overall difficulty scaling is statistically consistent, certain seemingly simple reasoning requirements remain more challenging for LLM agents than human experts initially anticipated during the manual annotation process.

\begin{figure}
    \centering
    \includegraphics[width=1\linewidth]{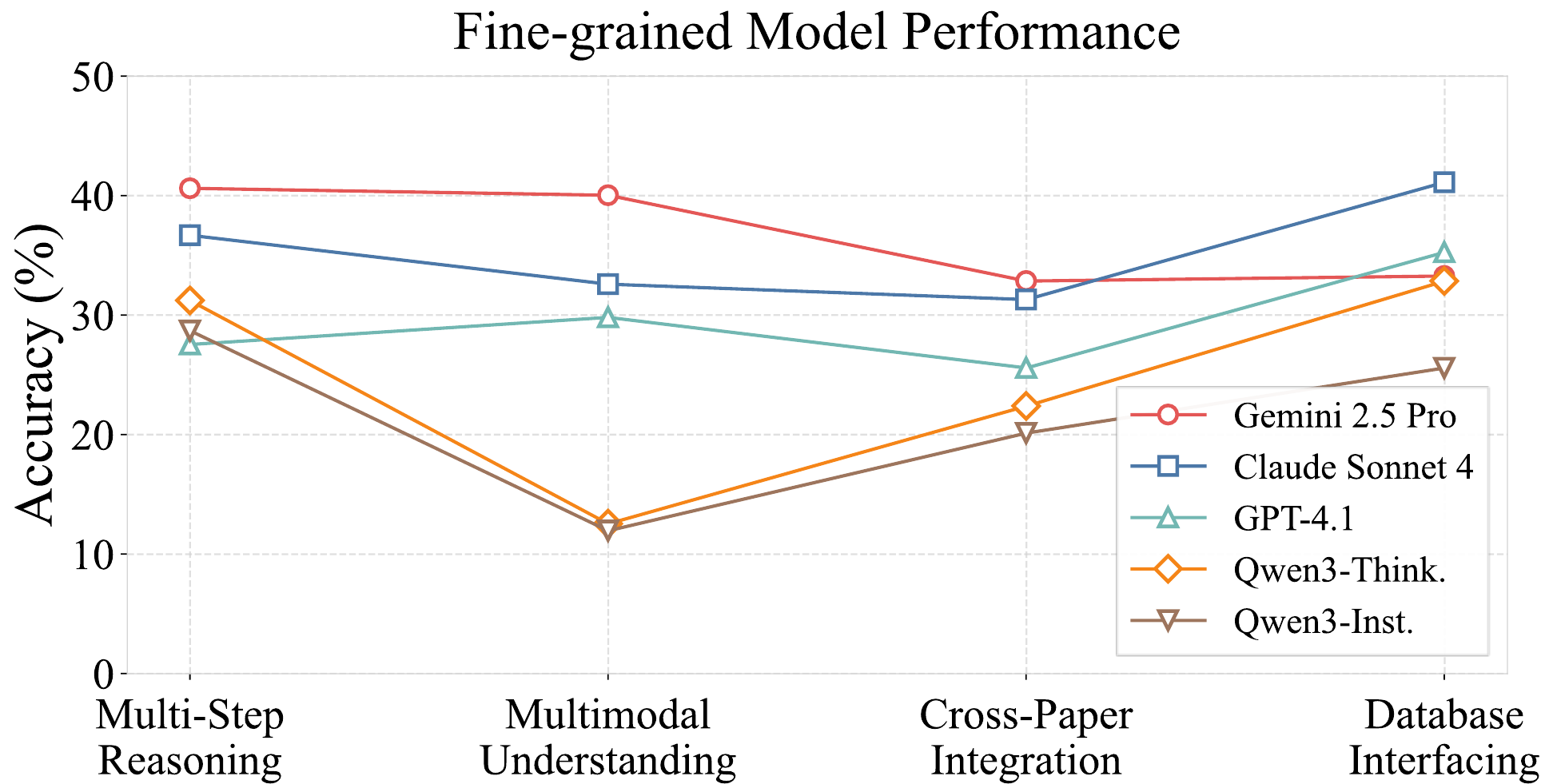}
    \caption{Fine-grained performance of five leading LLMs on the single-agent workflow, revealing their strengths and weaknesses across the four primary capabilities required in \Benchmark.}
    \label{fig:fig_9}
\end{figure}

\paragraph{Fine-Grained Agent Performance.}
To further probe the specific strengths and weaknesses of the top-performing models, we analyze the performance across the four primary capabilities of \Benchmark, as shown in Figure~\ref{fig:fig_9}: multi-step reasoning, multimodal understanding, cross-paper integration, and database interfacing. The performance breakdown, presented in the figure, reveals that no single LLM universally dominates across all capabilities. Gemini 2.5 Pro demonstrates superior proficiency in multi-step reasoning and multimodal understanding, showcasing its strengths in logic-heavy and visually dependent tasks. In contrast, Claude Sonnet 4 emerges as the standout performer in database interfacing, surpassing its competitors in these code-centric tasks. Meanwhile, both Qwen3 variants (Think and Inst.) exhibit a significant performance drop in multimodal understanding. This is because they are pure language models lacking intrinsic visual capabilities, necessitating reliance on external tools for image summarization. This limitation underscores the critical importance of native multimodal capabilities for autonomous agents. This fine-grained analysis highlights the often complementary strengths of leading LLMs. This insight suggests that a promising direction is to strategically leverage the distinct advantages of different specialized LLMs for appropriate sub-tasks, rather than relying on a single one.

\begin{figure*}[t]
    \centering
    \includegraphics[width=1\linewidth]{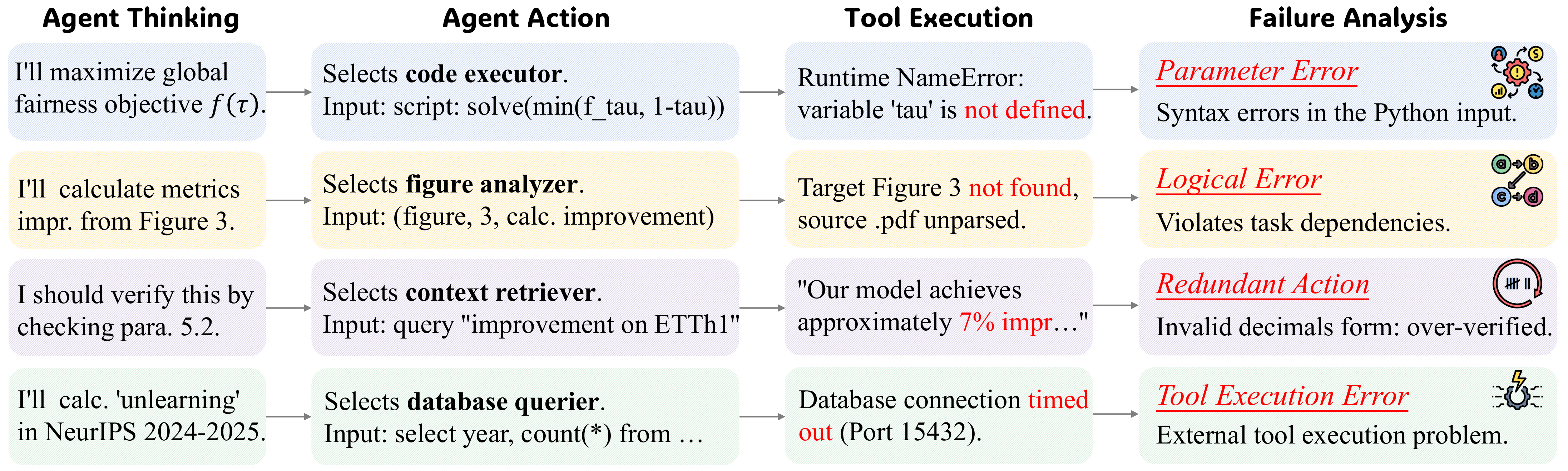}
    \caption{Failure analysis across the single-agent workflow, tracing the path from thinking to tool execution. This figure classifies failure modes into four distinct categories and illustrates specific instances for each: parameter errors such as syntax issues, logical errors like dependency violations, redundant actions such as over-verification, and tool execution errors like connection timeouts.}
    \label{fig:fig_10}
\end{figure*}

\paragraph{Failure Analysis.}
We conduct a failure attribution analysis of the agent reasoning process. We perform this experiment by inspecting tool outputs error messages and subsequent agent actions to attribute each failure. As shown in Figure \ref{fig:fig_10}, we categorize these issues into four types: parameter errors such as syntax faults in the input, logical errors like task dependency violations, redundant actions exemplified by over-verification, and tool execution errors stemming from external problems like connection timeouts. Our statistics indicate that tool execution errors account for less than 15\% of observed failures, suggesting the underlying tools remain reliable and do not constitute the primary performance bottleneck. Moreover, open-source LLMs exhibit more parameter and logical errors, indicating weaknesses in complex planning. In contrast, closed-source thinking LLMs tend toward redundant actions, suggesting a form of overthinking where agents needlessly re-verify known information. To mitigate these errors, agents could incorporate explicit memory modules that record the reasoning chain to guide subsequent actions and prevent repetition. Enhancing self-correction workflows and comprehensive error recovery would increase overall task success. The complete distribution of failures is detailed in Appendix \ref{sec:c}.

\subsection{Analysis of Evaluator Alignment}

\paragraph{Annotator Agreement.}
We established a reliable ground truth through a rigorous human verification process. A team of five Ph.D. experts refined the preliminary drafts by filtering trivial entries and balancing tool usage frequencies. The experts removed obvious indicators to maintain sufficient difficulty, and rewrote questions into authentic research formats. Subsequently, the annotators labeled each sample with the ground truth answer and the minimal viable tool chain, alongside the difficulty level and question type. To validate the reliability of this process, three independent experts evaluated a random subset of the data. The inter-annotator agreement (IAA) yields 0.83, which demonstrates high consistency among human evaluators.

\paragraph{Alignment of LLM and Human Judgement.}
We subsequently evaluate the consistency of GPT-4o as an automatic evaluator by comparing binary correctness labels against independent human scoring on the \Benchmark-Human subset. We employ both simple agreement rates and the Cohen Kappa to measure inter-rater reliability, while correcting for chance agreement~\cite{sun2011kappa}. The analysis reveals significant alignment between the model and human evaluators. We observe an agreement rate of 98.5\% and a Kappa of 0.97 for responses from Gemini 2.5 Pro. Similarly, the metrics for answers provided by Ph.D. experts are 98.0\% and 0.93, respectively. These values exceed the 0.8 threshold, which typically denotes near perfect agreement. The low number of discrepancies suggests that GPT-4o functions as a reliable proxy for assessing agent performance in this domain. The detailed calculation of metrics is provided in Appendix \ref{sec:d}.

\section{Conclusion and Future Work}

We propose \Benchmark, a novel evaluation benchmark tailored for tool-augmented reasoning over scientific literature. Unlike existing QA benchmarks that focus on tool-
free tasks within single papers, \Benchmark~requires multi-step, multimodal, cross-paper, database-aware reasoning grounded in authentic research scenarios. Experimental results quantify the limitations of current LLM-based agent workflows, showing that performance remains far below expert levels with an average accuracy of 38.78\%, dropping to merely 18.47\% on the hard subset. Beyond these quantitative metrics, we conduct both an analysis of reasoning traces and a diagnosis of agent behavior. These investigations reveal that performance bottlenecks stem from various aspects like inefficient tool usage and logic failure, providing critical insights into the fragility of existing agents in addressing research questions in scientific literature.

\Benchmark~and \Platform~facilitate the evaluation of novel strategies and the testing of custom tools. Future directions include enhancing the question generation pipeline and expanding the scope to broader scientific disciplines alongside new metrics for interpretability. We invite the community to utilize this framework to advance the development of capable agents for scientific discovery.

\newpage

\section*{Impact Statement}
This paper contributes to application-driven machine learning by evaluating the practical utility of tool-augmented agents in scientific research. \Benchmark\ serves as a testbed for assessing the reliability and precision essential for real-world deployment. By highlighting current limitations in agentic reasoning and tool usage, our work supports the development of trustworthy AI assistants that can eventually enhance research productivity across scientific disciplines.

Regarding ethical considerations, all source papers were retrieved from public, open-access platforms via unbiased, automated sampling. To ensure data integrity and mitigate potential biases, both the dataset construction and the experimental evaluation were conducted in an anonymized manner, with all author and institutional information strictly excluded. The resulting dataset is intended solely for non-commercial academic research, with all generated artifacts released under an open license to foster reproducibility. Any commercial exploitation is strictly prohibited.

\bibliography{example_paper}
\bibliographystyle{icml2026}

\newpage
\appendix
\onecolumn

\section{Benchmark Construction Details}
\label{sec:a}

In this section, we provide a granular breakdown of the construction process for the \Benchmark. We detail the preparation of the paper corpus, the hierarchical sampling strategy, and the question generation pipeline involving human experts.

\subsection{Paper Corpus Preparation}

The foundation of the \Benchmark\ benchmark is a large-scale corpus of academic papers. We adhered to a rigorous data processing pipeline consisting of collection, representation, and sampling stages to ensure validity and timeliness.

\paragraph{Data Collection and Preprocessing.}
We performed a comprehensive crawl of open-access repositories, specifically targeting OpenReview and Open Access platforms. To minimize data contamination where models might have memorized content during pre-training, we strictly filtered for papers published in the year 2025. The raw collection resulted in 14,435 papers. We subsequently removed papers with incomplete metadata, non-English content, or those limited to abstract-only availability to ensure the quality of the source material.

\paragraph{Feature Representation.}
To analyze the distribution of the corpus, we mapped each paper to a semantic feature vector. We employed GPT-4o-mini as an annotator to extract key attributes from the abstract and introduction of each paper. These attributes are encoded into a 20-dimensional binary vector structured across four primary categories. The research field category covers eight domains: machine learning (ML), natural language processing (NLP), computer vision (CV), reinforcement learning (RL), generative AI (GenAI), information retrieval (IR), AI for science (AI4Science), and machine learning systems (MLSys). The remaining dimensions encode the methodology category, such as theoretical analysis or architecture design; the evaluation paradigm, including human evaluation or standard benchmarks; and proxy metrics for potential impact. We visualized these vectors using t-SNE with a perplexity of 30 and 1,000 iterations. As shown in Figure~\ref{fig:fig_2}(a), the visualization reveals dense clusters representing mainstream research topics, highlighting the necessity for a balanced sampling strategy.

\paragraph{Hierarchical Sampling Strategy.}
To construct a benchmark of 100 papers that balances representativeness and diversity, we designed a two-stage hierarchical sampling method.

We first aimed to identify papers that act as prototypes for the dominant research clusters. We applied the K-Medoids algorithm to select 50 centers. Unlike K-Means, K-Medoids selects actual data points as centers, ensuring the chosen papers are real and valid. The objective is to minimize the intra-cluster distance:
\begin{equation}
\mathcal{P}_{\text{prototype}} = \underset{\{P_{c_1}, \ldots, P_{c_K}\} \subset \mathcal{P}}{\arg\min} \sum_{i=1}^{N} \min_{k \in \{1,\ldots,K\}} d(\mathbf{v}_i, \mathbf{v}_{c_k}),
\end{equation}
where $d$ denotes the Hamming distance suitable for the binary feature vectors.

\begin{figure*}[t]
    \centering
    \includegraphics[width=1\linewidth]{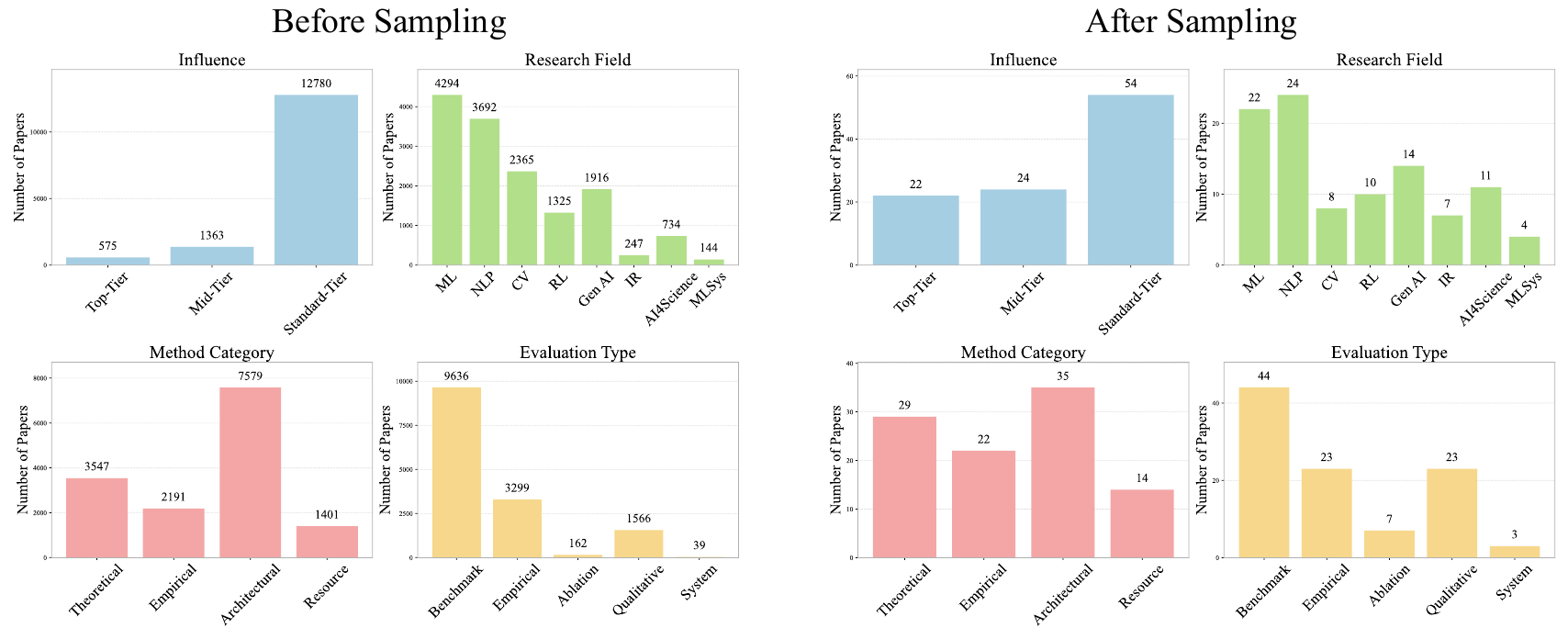}
    \caption{Comparison of paper distributions before and after sampling across four dimensions (influence, research field, method category, evaluation type). The sampled subset remains representative of the full corpus while preserving diversity across all categories.}
    \label{fig:fig_11}
\end{figure*}

To avoid evaluation bias where models are only tested on popular topics, we selected 50 boundary papers from the sparse regions of the feature space. We employed an iterative Farthest Point Sampling (FPS) strategy. At each step, we selected the paper that maximizes the minimum distance to the current selected set:
\begin{equation}
P_{b_i} = \underset{P_j \in \mathcal{P} \setminus \mathcal{Q}_{i-1}}{\arg\max} \left( \min_{P_s \in \mathcal{Q}_{i-1}} d(\mathbf{v}_j, \mathbf{v}_s) \right).
\end{equation}
The final subset comprises 100 papers. As illustrated in Figure~\ref{fig:fig_11}, this strategy shifts the distribution from being heavily skewed towards dominant fields and standard methods to a more uniform coverage across influence levels, research fields, methodologies, and evaluation types.

\subsection{QA Generation Pipeline}

Transforming the selected papers into high-quality QA pairs involved a sophisticated pipeline that combines automated parsing, LLM generation, and rigorous expert verification.

\paragraph{Definition of the Toolset.}

We define a comprehensive toolset to simulate a realistic agentic workflow, ensuring that the agent has sufficient capabilities to address complex queries while operating within clear constraints. The comprehensive toolset $\mathcal{T}$ is formally defined as the union of the following specialized modules:
\begin{equation}
    \mathcal{T} = \{\mathcal{T}_{\text{pdf}}, \mathcal{T}_{\text{tab}}, \mathcal{T}_{\text{fig}}, \mathcal{T}_{\text{ref}}, \mathcal{T}_{\text{web}}, \mathcal{T}_{\text{rag}}, \mathcal{T}_{\text{db}}, \mathcal{T}_{\text{code}}\}.
\end{equation}
Here, $\mathcal{T}_{\text{pdf}}$, $\mathcal{T}_{\text{tab}}$, and $\mathcal{T}_{\text{fig}}$ denote tools for parsing PDF documents, tables, and figures, respectively. $\mathcal{T}_{\text{ref}}$ handles reference management, while $\mathcal{T}_{\text{web}}$ and $\mathcal{T}_{\text{rag}}$ provide access to external web content and retrieval-augmented generation pipelines. Finally, $\mathcal{T}_{\text{db}}$ and $\mathcal{T}_{\text{code}}$ allow for structured database queries and executable code generation.

\paragraph{Structure-Aware Parsing.}
Standard text extraction often discards structural cues and visual information vital for understanding scientific papers. We utilized MinerU2.5~\cite{wang2024mineru} to convert the raw PDF of each paper into a structured Markdown format. Crucially, this process goes beyond text: it extracts and crops figures and charts into high-resolution image files to support multimodal analysis, and explicitly parses references into structured citation data. By preserving hierarchical headers, block equations, visual assets, and citation links, we establish a comprehensive grounding context for high-fidelity question generation.

\paragraph{Multi-Model Draft Generation.}
Relying on a single model for question generation can introduce specific cognitive biases. To ensure diversity in reasoning patterns, we prompted three distinct frontier LLMs: GPT-5~\cite{singh2025openai}, Claude Sonnet 4~\cite{Anthropic2025Claude4}, and Gemini 2.5 Pro~\cite{comanici2025gemini}. Acting as independent question proposers, these models were fed the structured paper segments and the precise definitions of the toolset $\mathcal{T}$. We employed prompt \ref{pro:1} to enforce constraints that required questions to necessitate multi-hop reasoning and tool execution, specifically avoiding simple lookup queries. This ensemble approach leverages the unique strengths of each model architecture, yielding a robust and diverse pool of 2,875 initial drafts.

\paragraph{Human-in-the-Loop Refinement.}
The raw drafts generated by LLMs often exhibit triviality, hallucinated content, or rigid formatting that fails to mimic real-world inquiries. To bridge this gap, we assembled a team of domain experts with Ph.D. degrees to conduct a labor-intensive, three-pass refinement process. In the initial filtration phase, experts manually scrutinized thousands of drafts, ruthlessly discarding entries that were answerable via the abstract alone, logically flawed, or factually unsupported by the source text. Subsequently, in the rewriting phase, the team transformed the remaining candidates from mechanical, pointer-heavy instructions (e.g., "According to Table 3...") into authentic, intent-driven research questions. This step required significant cognitive effort to ensure questions remained solvable while removing direct clues, thereby compelling agents to perform autonomous information retrieval. Finally, during the annotation and verification phase, experts rigorously validated the ground truth for each question by tracing the exact evidence within the paper. They further analyzed the problem-solving logic to define the minimal viable tool chain and calibrated the difficulty level (Easy, Medium, Hard) based on the depth of reasoning required. This comprehensive human-in-the-loop workflow ensures the benchmark upholds the highest standards of scientific rigor and complexity.

\subsection{Benchmark Statistics}

\begin{wraptable}{r}{0.55\textwidth}
    \centering
    \small
    \caption{Distribution of question types and difficulty levels in \Benchmark~and \Benchmark-Human. MC: Multiple Choice, CA: Concise Answer, OA: Open Answer.}
    \label{tab:tab_3}
    \begin{tabular}{l ccc | c | ccc | c}
        \toprule
        \multirow{2}{*}{\textbf{Difficulty}} & \multicolumn{4}{c|}{\textbf{\Benchmark}} & \multicolumn{4}{c}{\textbf{\Benchmark-Human}} \\
        & MC & CA & OA & Total & MC & CA & OA & Total \\
        \midrule
        Easy   & 38 & 106 & 23  & 167 & 8  & 24 & 8   & 40  \\
        Medium & 61 & 185 & 122 & 368 & 19 & 48 & 30  & 97  \\
        Hard   & 27 & 82  & 140 & 249 & 7  & 22 & 34  & 63  \\
        \midrule
        Total  & 126 & 373 & 285 & 784 & 34 & 94 & 72  & 200 \\
        \bottomrule
    \end{tabular}
\end{wraptable}

\paragraph{Question Type and Difficulty.}
The benchmark includes three question formats: Multiple Choice (MC), Concise Answer (CA), and Open Answer (OA). Each question is assigned a difficulty level of Easy, Medium, or Hard. Table~\ref{tab:tab_3} shows the distribution for both the full benchmark, which we call \Benchmark~(784 questions), and a 200-question subset reserved for human evaluation, \Benchmark-Human. The distribution highlights a focus on Medium difficulty questions and a substantial number of open-ended questions to rigorously test generative capabilities.

\begin{wraptable}{r}{0.55\textwidth}
    \centering
    \footnotesize
    \caption{Distribution of questions across key features in the \Benchmark.}
    \label{tab:tab_4}
        \begin{tabular}{ccccc}
            \toprule
            \textbf{\begin{tabular}[c]{@{}c@{}}Multi-Step \\ Reasoning\end{tabular}} & 
            \textbf{\begin{tabular}[c]{@{}c@{}}Multimodal \\ Understanding\end{tabular}} & 
            \textbf{\begin{tabular}[c]{@{}c@{}}Cross-paper \\ Integration\end{tabular}} & 
            \textbf{\begin{tabular}[c]{@{}c@{}}Database \\ Interfacing\end{tabular}} & 
            \textbf{Total} \\
            \midrule
            170 & 261 & 194 & 159 & 784 \\
            \bottomrule
        \end{tabular}
\end{wraptable}

\paragraph{Key Features.}
We categorize the questions into four primary dimensions as illustrated in Table~\ref{tab:tab_4}: Multi-Step Reasoning, Multimodal Understanding, Cross-Paper Integration, and Database Interfacing. The data distribution across these categories is notably balanced, with each feature represented by a substantial yet focused sample size. This well-proportioned allocation ensures that the evaluation remains comprehensive across all critical research tasks without being skewed toward any single dimension, thereby maintaining both the breadth and the rigor of the benchmark.

\section{Evaluation Details on \Platform}
\label{sec:b}

\subsection{Tool Environment Implementation}
We implement a comprehensive tool suite on \Platform\ to facilitate interaction with scientific literature and external knowledge sources. Table~\ref{tab:tab_8} presents the specific tool inputs and parameters. To ensure security and reproducibility, we deploy all computational tools within isolated Docker containers. These environments enforce strict constraints on network access and computing resources to prevent malicious operations. Additionally, the execution layer features a robust error handling mechanism. Upon failure, the system captures standard error output and returns a structured message, which enables the agent to debug and self correct within the allocated reasoning steps.

\subsection{Evaluation Setup}
\paragraph{Agent Baselines.} We evaluate nine state-of-the-art LLMs: Gemini-2.5-Pro-2506, OpenAI-o4-mini-high-2504, Claude-Sonnet-4-2505, GPT-4.1-2504, Claude-3.5-Sonnet-2410, Qwen3-235B-Thinking-2507, GLM-4.5-2507, Qwen3-235B-Instruct-2507, and Kimi-K2-Instruct-2507. For all agent-based evaluations, we use a consistent set of inference parameters: a temperature of 0.7, top\_p of 0.9, a maximum of 4096 output tokens, and a maximum of 40 reasoning steps per question. Additionally, we employ a standardized chain-of-thought (CoT) \cite{wei2022chain} prompting strategy. The system prompt instructs the agent to think step-by-step, explicitly stating the rationale before invoking any tool. We conduct three independent runs for each experimental setting.

\paragraph{Human Baselines.} To establish a human expert baseline, we recruit three computer science Ph.D. candidates to complete the \Benchmark-Human~subset. The evaluation is conducted under controlled conditions: participants are allotted 16 hours over two days and are permitted to use standard tools, including GPT-4o for document comprehension, web search for information retrieval, and Python for coding tasks.

\begin{table*}[t!]
    \centering
    \caption{Detailed specification of the implemented tool library on the \Platform.}
    \label{tab:tab_8}
    \resizebox{\textwidth}{!}{%
        \begin{tabular}{p{2.5cm} p{4.5cm} p{3cm} p{4cm} p{2.5cm} p{5cm}}
        \toprule
        \textbf{Tool Name} & \textbf{Description} & \textbf{Input} & \textbf{Output} & \textbf{Parameters} & \textbf{Detailed Operation} \\
        \midrule
        \makecell[tl]{\textbf{PDF} \\ \textbf{Parser}} & Parses the non-readable PDF into an LLM-readable structured format. & PDF binary file & Markdown content and each layout block (JSON). & \texttt{pdf\_path} & Employs the MinerU to extract text, tables, and figures along with their layout. \\
        \addlinespace
        \makecell[tl]{\textbf{Database} \\ \textbf{Querier}} & Retrieves papers from a vector database based on topic or keywords. & Conference name, paper status (e.g., Poster). & List of relevant paper information. & \texttt{conference}, \texttt{status} & First applying BM25 for sparse retrieval, followed by dense retrieval using cosine similarity. \\
        \addlinespace
        \makecell[tl]{\textbf{Code} \\ \textbf{Executor}} & Executes code snippets in a sandboxed environment for calculations or replication. & String containing code to be executed. & Standard output (stdout) and standard error (stderr) & \texttt{code}, \texttt{timeout} & Utilizes a secure, sandboxed Python environment to execute arbitrary code strings and return the execution results. \\
        \addlinespace
        \makecell[tl]{\textbf{Cross-Ref} \\ \textbf{Searcher}} & Finds and resolves bibliographic citations within a paper using arxiv APIs. & Citation key string from the paper text. & Structured citation data including title, abstract, and URL. & \texttt{citation\_key} & Extracts reference text from the document and leverages GPT-4o to parse and retrieve corresponding bibliographic details. \\
        \addlinespace
        \makecell[tl]{\textbf{Web} \\ \textbf{Searcher}} & Queries web search engines to find background information on concepts, terms, or techniques. & Search query string. & Snippets of information and corresponding source URLs. & \texttt{query}, \texttt{n\_nums} & Interfaces with the SerpAPI to perform web searches and parse the returned results into a structured format. \\
        \addlinespace
        \makecell[tl]{\textbf{Context} \\ \textbf{Retriever}} & Performs semantic search over long texts to find relevant passages~\cite{yu2024multi}. & Query string. & Relevant text chunks. & \texttt{query}, \texttt{top\_k} & Uses \texttt{text-embedding-3-small} to embed the query and chunks, then performs similarity search. \\
        \addlinespace
        \makecell[tl]{\textbf{Figure} \\ \textbf{Analyzer}} & Analyzes visual content within figures and charts to extract information. & A specific question about the figure. & Extracted textual or numerical data from the figure. & \texttt{figure\_id}, \texttt{query} & Feeds the figure and its caption directly to a multimodal LLM. If the base LLM is not multimodal, it first enriches the caption via GPT-4o. \\
        \addlinespace
        \makecell[tl]{\textbf{Table} \\ \textbf{Analyzer}} & Extracts specific data points or verifies trends from tables by parsing their structure. & A specific question about the table. & Extracted textual or numerical data from the table. & \texttt{table\_id}, \texttt{query} & Converts the table image into structured HTML for LLM processing to avoid context limits of image input. \\
        \bottomrule
    \end{tabular}%
    }
\end{table*}

\subsection{Evaluation Metric Definitions}
We assess performance using three primary metrics and a supplementary item response theory (IRT) analysis.

\paragraph{Accuracy.} We employ the LLM-as-a-Judge where GPT-4o assigns a binary score $S_{\text{corr}} \in \{0, 1\}$ to each answer. An answer is deemed correct if its final conclusion is accurate and supported by valid reasoning, regardless of stylistic format.

\paragraph{Reasoning Steps.} To measure solution complexity, we compute the average number of tool calls in an agent's executed tool chain, $\tau^{\text{exec}}$. For a set of questions $Q$, the average reasoning steps are:
\begin{equation}
    \text{Reasoning Steps} = \frac{1}{|Q|} \sum_{q \in Q} |\tau^{\text{exec}}_q|.
\end{equation}

\paragraph{Efficiency.} To assess how concisely an agent solves a problem, we measure the overlap between the executed tool chain $\tau^{\text{exec}}$ and the ground-truth chain $\tau^{\text{gt}}$, normalized by the length of the executed chain:
\begin{equation}
    \text{Efficiency} = \frac{1}{|Q|} \sum_{q \in Q} \frac{|\tau^{\text{exec}}_q \cap \tau^{\text{gt}}_q|}{|\tau^{\text{exec}}_q|}.
\end{equation}
To ensure reliability, we manually verify all human answers and LLM responses on the \Benchmark-Human~subset.

\paragraph{IRT Analysis.}
To disentangle the intrinsic ability of the models from the difficulty of the benchmark questions, we employ a 2-parameter logistic (2PL) item response theory model. This allows us to estimate the probability of a model $j$ correctly answering question $i$. The probability $P_{ij}(\theta_j)$ is defined as:

\begin{equation}
    P_{ij}(\theta_j) = \frac{1}{1 + e^{-a_i(\theta_j - \beta_i)}},
\end{equation}

where $\theta_j$ represents the latent ability of model $j$ (higher $\theta$ indicates a more capable model); $\beta_i$ denotes the difficulty parameter of question $i$, representing the point on the ability scale where an agent has a 0.5 probability of answering correctly; and $a_i$ is the discrimination parameter of question $i$, indicating how well the question differentiates between models of varying abilities (a higher $a_i$ implies a steeper slope at the inflection point). We estimate parameters using marginal maximum likelihood (MML) estimation with markov chain monte carlo (MCMC) sampling. This analysis verifies that \Benchmark~maintains a balanced difficulty distribution and effectively distinguishes between models of similar performance.

\section{Detailed Results}
\label{sec:c}

\begin{wraptable}{r}{0.55\textwidth}
    \centering
        \caption{Fine-grained performance of five leading LLMs on the single-agent workflow, revealing their specialized strengths and weaknesses across the four primary features in questions on \Benchmark.}
        \label{tab:tab_6}
        \resizebox{\linewidth}{!}{
            \begin{tabular}{lcccc}
                \toprule
                \textbf{Base LLM} & 
                \textbf{\begin{tabular}[c]{@{}c@{}}Multi-Step \\ Reasoning\end{tabular}} & 
                \textbf{\begin{tabular}[c]{@{}c@{}}Multimodal \\ Understanding\end{tabular}} & 
                \textbf{\begin{tabular}[c]{@{}c@{}}Cross-paper \\ Integration\end{tabular}} & 
                \textbf{\begin{tabular}[c]{@{}c@{}}Database \\ Interfacing\end{tabular}} \\
                \midrule
                Gemini 2.5 Pro & 32.85 & 40.03 & 40.61 & 33.27 \\
                Claude Sonnet 4 & 31.30 & 32.59 & 36.68 & 41.12 \\
                GPT-4.1 & 25.58 & 29.81 & 27.54 & 35.26 \\
                Qwen3-Think. & 22.39 & 12.54 & 31.22 & 32.85 \\
                Qwen3-Inst. & 20.13 & 11.96 & 28.70 & 25.58 \\
                \bottomrule
            \end{tabular}
        }
\end{wraptable}

\paragraph{Fine-grained Agent Performance.} Table~\ref{tab:tab_6} presents the performance of five large language models across four capabilities. The data indicates that Gemini 2.5 Pro achieves the highest proficiency in multi-step reasoning and multimodal understanding. Claude Sonnet 4 demonstrates superior capability in database interfacing and outperforms others in code-centric interactions. The results also confirm the significant deficit of Qwen3 variants in multimodal understanding due to the lack of native visual processing. This breakdown reinforces that different models possess complementary advantages suitable for distinct sub-tasks.

\begin{wrapfigure}{r}{0.55\textwidth}
    \centering
    \includegraphics[width=\linewidth]{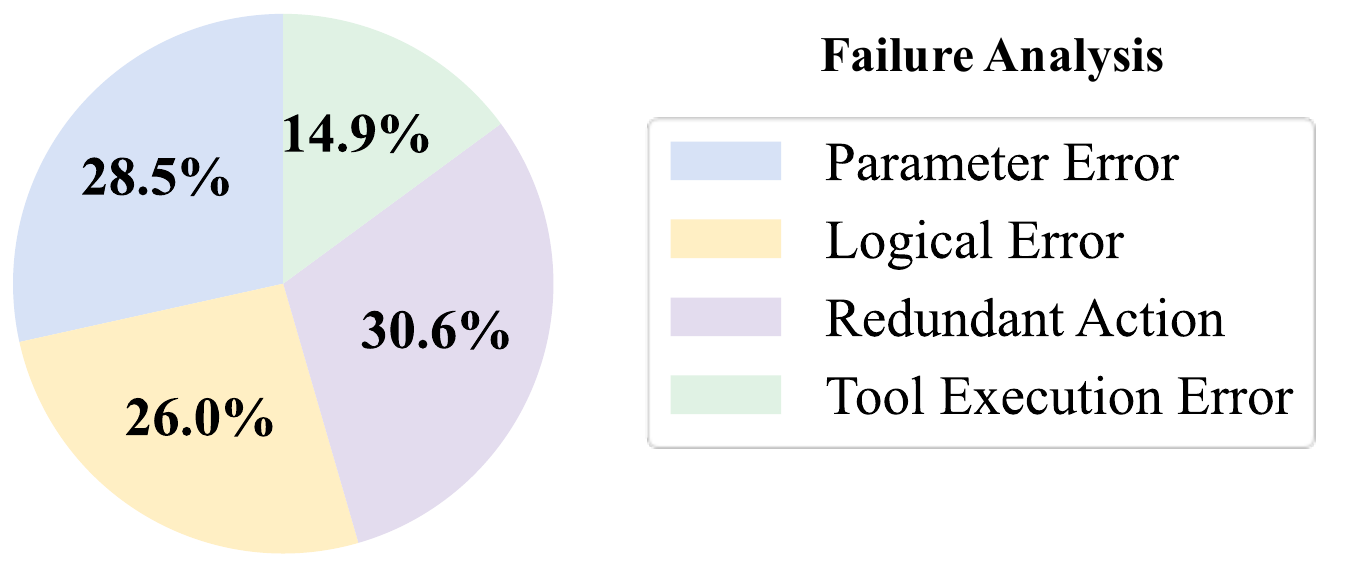}
    \caption{Failure analysis of five leading LLMs on the single-agent workflow, presenting the proportion of each error type in observed failures.
    }
    \label{fig:fig_12}
\end{wrapfigure}

\paragraph{Failure Analysis.} Figure~\ref{fig:fig_12} illustrates a comprehensive failure attribution analysis regarding the agent reasoning process. We performed this experiment by systematically inspecting tool output error messages and subsequent agent actions to accurately attribute each failure instance. These quantitative results are derived by averaging the error rates across the five models listed in Table~\ref{tab:tab_6}. Crucially, our statistics indicate that tool execution errors account for less than 15\% of observed failures. This finding attests to the robustness of our platform, suggesting that the underlying tools remain highly reliable and do not constitute the primary performance bottleneck.

\begin{wrapfigure}{r}{0.55\textwidth}
    \centering
    \includegraphics[width=\linewidth]{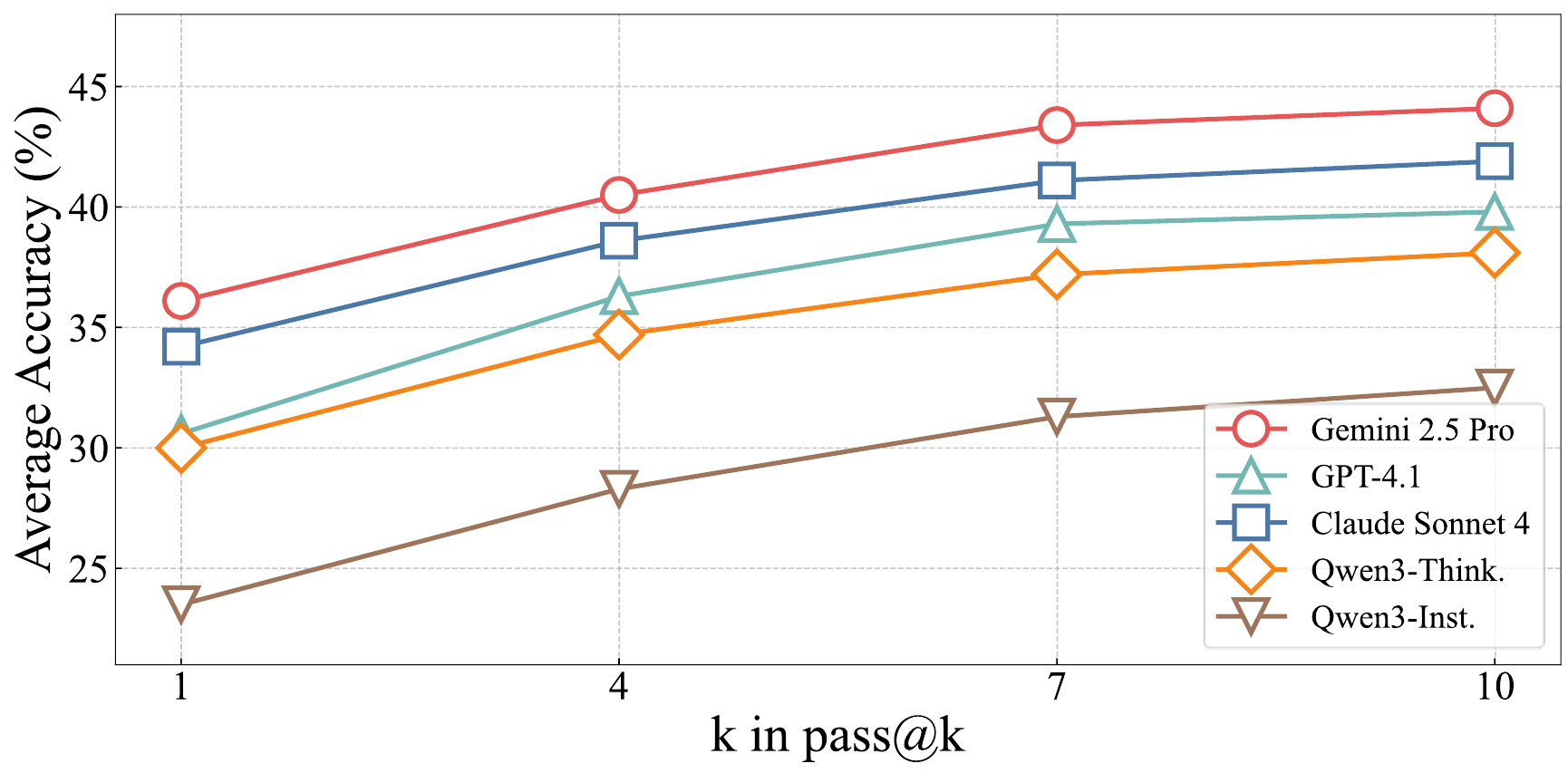}
    \caption{Performance in the pass@$k$ evaluation reveals limits of test-time scaling capabilities on \Benchmark. By trying multiple times and taking consistent answers, the agent achieves better performance.
    }
    \label{fig:fig_13}
\end{wrapfigure}

\paragraph{Pass@$k$ Performance.}

We evaluate agent test-time scaling in a pass@$k$ scenario, where each task is executed $k$ times under identical parameters and the result is determined by a majority vote. Figure~\ref{fig:fig_13} reveals a clear positive correlation between attempts and agent performance, a trend consistent across all models. Notably, this scaling effect allows fast-thinking models to match or surpass the single-pass performance of slow-thinking ones. However, this approach introduces a significant challenge: a naive parallel implementation leads to nearly linear cost growth with respect to $k$. Compounding this issue, we find that performance gains exhibit diminishing returns; the improvement is substantial for small $k$ but becomes marginal as $k$ increases. These dual findings highlight the central problem of achieving an optimal trade-off between performance gains and evaluation cost. Future work can focus on designing agents with larger token budgets under controlled cost constraints and developing more efficient parallelization strategies that move beyond simple majority voting.

\section{Human Annotation and Evaluation Details}
\label{sec:d}

\subsection{Human Annotation Guideline}
To ensure the high quality and difficulty of the benchmark, we employed a Human-in-the-Loop refinement process following the human annotator instruction \ref{pro:2}, where a team of five Ph.D. experts transformed preliminary LLM-generated drafts into authentic, solvable, and challenging research questions. To strictly validate the reliability of this process, three independent experts subsequently evaluated a random subset of the data. We quantified the consistency among these three evaluators using Krippendorff's alpha ($\alpha$), calculated as:
\begin{equation}
\alpha = 1 - \frac{D_o}{D_e}.
\end{equation}
In this formula, $D_o$ represents the observed disagreement derived from weighted discrepancies among the three raters, while $D_e$ denotes the disagreement expected by chance. The resulting $\alpha$ of 0.83 demonstrates high consistency. The comprehensive guidelines provided to the annotators are presented below.

\subsection{Calculation of Alignment Metrics}

To rigorously quantify the consistency between annotators (for the ground truth dataset) and between the LLM evaluator and human experts, we utilized two primary metrics: Simple Agreement Rate and Cohen Kappa.

\paragraph{Simple Agreement Rate.}
The Simple Agreement Rate (denoted as $p_o$) measures the raw proportion of instances where two evaluators assign the identical label (e.g., Correct or Incorrect) to a given response. It is calculated as:

\begin{equation}
    p_o = \frac{1}{N} \sum_{i=1}^{N} \mathbb{I}(y_i^{(A)} = y_i^{(B)}),
\end{equation}

\noindent where $N$ is the total number of samples, $y_i^{(A)}$ and $y_i^{(B)}$ represent the binary labels assigned by evaluator A and evaluator B for the $i$-th sample, respectively. $\mathbb{I}(\cdot)$ is the indicator function, which equals 1 if the condition inside is true and 0 otherwise.

\paragraph{Cohen Kappa.}
While simple agreement is intuitive, it does not account for the possibility of agreement occurring by random chance. To address this, we employ Cohen Kappa coefficient ($\kappa$)~\cite{sun2011kappa}, a robust statistic for inter-rater reliability. The coefficient is defined as:

\begin{equation}
    \kappa = \frac{p_o - p_e}{1 - p_e}.
\end{equation}

\noindent Here, $p_o$ corresponds to the observed agreement rate defined in Eq.~(1). The term $p_e$ represents the hypothetical probability of chance agreement, calculated based on the marginal distributions of the labels. For a binary classification task (e.g., Correct/Incorrect) with two evaluators, $p_e$ is computed as:

\begin{equation}
    p_e = \left( \frac{n_{C}^{(A)}}{N} \times \frac{n_{C}^{(B)}}{N} \right) + \left( \frac{n_{I}^{(A)}}{N} \times \frac{n_{I}^{(B)}}{N} \right),
\end{equation}

\noindent where $n_{C}^{(j)}$ and $n_{I}^{(j)}$ denote the count of samples labeled as correct and incorrect by evaluator $j$ (where $j \in \{A, B\}$), respectively. A $\kappa$ value greater than 0.8 is typically interpreted as indicating near-perfect agreement.

\begin{StrategyBox}[frameorange]{bgorange}{Agentic QA Generation Prompt}
\refstepcounter{idx}
\label{pro:1}
\noindent\textbf{1. ROLE} \\
You are a Senior AI Research Scientist designing a Benchmark for Agentic Reasoning. Your task is to generate \textbf{10 extremely challenging, multi-hop questions} based on the provided paper. \\

\noindent\textbf{CRITICAL CONSTRAINT}: \\
Every question must require a \textbf{complex tool-chain of 4 to 10 steps}. Questions must be impossible to answer by simply reading the provided PDF. They \textbf{must} require retrieving external papers, verifying official documentation, or performing code-based re-calculations. \\

\noindent\textbf{2. STRICT TOOL DEFINITIONS} \\
Your questions must be solvable \textit{only} by combining the following specific tools: \\
\noindent\texttt{PDF Parser}: Parses the content of a PDF file. \\
\noindent\texttt{Context Retriever}: Search and retrieve text chunks from a parsed document. \\
\noindent\texttt{Table Analyzer}: Extracts structured data from tables. \\
\noindent\texttt{Figure Analyzer}: Extracts visual information from charts/diagrams. \\
\noindent\texttt{Cross-Ref Searcher}: Finds metadata/links for citations (e.g., ``[12]'') mentioned in the text. \\
\noindent\texttt{Web Searcher}: Searches the live internet for documentation or general info. \\
\noindent\texttt{Database Querier}: Queries specific academic databases for paper lists (e.g., by conference). \\
\noindent\texttt{Code Executor}: Runs Python code for calculation or algorithm verification. \\

\noindent\textbf{3. QUESTION TEMPLATES} \\
You can use the following templates as reference: \\
\noindent\textbf{Templates 1} \\
\textbf{Chain:} \texttt{PDF Parser} $\rightarrow$ \texttt{Table Analyzer} $\rightarrow$ \texttt{Cross-Ref Searcher} $\rightarrow$ \texttt{PDF Parser} $\rightarrow$ \texttt{Table Analyzer} $\rightarrow$ \texttt{Code Executor} \\
\textbf{Example:} ``Table 2 reports the `Llama-2' baseline as 65.4. Retrieve the original Llama-2 paper referenced in Citation [12], find the reported score on this \textbf{exact same dataset split} in that paper's results table, and uses code to \textbf{calculate the precise numerical difference}.'' \\
\noindent\textbf{Templates 2} \\
\textbf{Chain:} \texttt{PDF Parser} $\rightarrow$ \texttt{Context Retriever} $\rightarrow$ \texttt{Cross-Ref Searcher} $\rightarrow$ \texttt{PDF Parser} $\rightarrow$ \texttt{Context Retriever} $\rightarrow$ \texttt{Code Executor} \\
\textbf{Example:} ``Equation (4) uses a weighting factor $\lambda$ which the text states is adopted from [Citation 3]. Find the value of $\lambda$ defined in [Citation 3], then use the raw inputs from Table 1 of the current paper to \textbf{re-calculate the loss result} via code. Does it match?'' \\
\noindent\textbf{Templates 3} \\
\textbf{Chain:} \texttt{PDF Parser} $\rightarrow$ \texttt{Figure Analyzer} $\rightarrow$ \texttt{Web Searcher} $\rightarrow$ \texttt{Context Retriever} \\
\textbf{Example:} ``Figure 3 illustrates the architecture using a `Multi-Head Attention' block. Search for the official PyTorch documentation for \texttt{nn.MultiheadAttention}. \textbf{Identify which input parameter} shown in Figure 3's diagram is NOT present in the standard PyTorch v2.0 API implementation.'' \\
\noindent\textbf{Templates 4} \\
\textbf{Chain:} \texttt{PDF Parser} $\rightarrow$ \texttt{Table Analyzer} $\rightarrow$ \texttt{Database Querier} $\rightarrow$ \texttt{PDF Parser} $\rightarrow$ \texttt{Table Analyzer} $\rightarrow$ \texttt{Code Executor} \\
\textbf{Example:} ``The paper claims SOTA performance on Dataset X compared to methods up to 2024. Use the database to find top-tier papers from CVPR 2025 that evaluate on Dataset X. \textbf{Compare the best score} from those 2025 papers against this paper's result and report the difference.'' \\

\end{StrategyBox}

\begin{StrategyBox}[frameblue]{bgblue}{Human Annotator Guideline}
\refstepcounter{idx}
\label{pro:2}
\noindent\textbf{1. ROLE} \\
You are a Senior Research Scientist and Benchmark Evaluator. Your task is to review, refine, and annotate question-answer pairs derived from scientific literature. \\

\noindent\textbf{2. OBJECTIVE} \\
Transform raw, LLM-generated draft questions into a high-quality benchmark dataset that tests the capabilities of autonomous research agents. \\

\noindent\textbf{PHASE 1: FILTERING \& REWRITING} \\

\noindent\textbf{Validity Check:} Discard questions that are hallucinated, trivial (lookup of a single keyword), or unsolvable even with the provided toolset $\mathcal{T}$. \\
\textbf{De-contextualization:} Remove phrases that reveal the source immediately. The question must simulate a genuine user query (e.g., ``What is the specific F1-score achieved by Method X in the dataset Y?''). \\
\textbf{Authenticity Refinement:} Rewrite the question to sound natural yet professional, ensuring clearly defined operational boundaries. \\

\noindent\textbf{PHASE 2: ANNOTATION \& LABELING} \\

\noindent\textbf{Ground Truth:} Provide the exact answer and specify the acceptable margin of error or unit for numerical values. \\
\textbf{Minimal Viable Tool Chain:} List the strictly necessary sequence of tools (e.g., \texttt{PDF\_Parser} $\rightarrow$ \texttt{Code\_Executor}) required for efficiency. \\
\textbf{Question Type:} Classify into \textit{Multiple Choice} (select best option), \textit{Concise Answer} (specific entity/number), or \textit{Open Answer} (synthesis/explanation). \\
\textbf{Difficulty Level:} Categorize as \textit{Easy} (direct retrieval), \textit{Medium} (multi-hop reasoning), or \textit{Hard} (complex synthesis, code execution, or cross-referencing).
\end{StrategyBox}


\end{document}